\documentclass[format=acmsmall, nonacm=true, review=false, screen=true]{acmart}

\usepackage{geometry}

\usepackage[utf8]{inputenc}

\usepackage{booktabs} % For formal tables

\usepackage[ruled]{algorithm2e} % For algorithms

\usepackage{pifont}

\usepackage{todonotes}  % For internal discussions
\usepackage{glossaries} % For abbreviations

\usepackage{array, longtable}
\newcolumntype{L}[1]{>{\raggedright\let\newline\\\arraybackslash\hspace{0pt}}m{#1}}
\newcolumntype{C}[1]{>{\centering\let\newline\\\arraybackslash\hspace{0pt}}m{#1}}
\newcolumntype{R}[1]{>{\raggedleft\let\newline\\\arraybackslash\hspace{0pt}}m{#1}}

\newcommand{\smltick}{\footnotesize{\checkmark}}

\usepackage{enumerate}
\usepackage{xcolor}

\usepackage{color, colortbl}

\usepackage[normalem]{ulem}

\usepackage{multirow}
\usepackage{marginnote}

\usepackage{tikz}
\usetikzlibrary{arrows,shapes,positioning,shadows,trees}
\tikzset{
  basic/.style  = {draw, text width=2cm, drop shadow, font=\sffamily, rectangle, fill=white},
  root/.style   = {basic, rounded corners=2pt, thin, align=center},
  level 2/.style = {basic, rounded corners=6pt, thin, align=center, text width=8em},
  level 3/.style = {basic, thin, align=left, , text width=6.5em}
}

\newacronym{aaip}{AAIP}{Assuring Autonomy International Programme}

\newacronym{bit}{BIT}{Built-In Test}

\newacronym{cbar}{C-BAR}{Critical Barrier}

\newacronym{eda}{EDA}{Exploratory Data Analysis}
\newacronym{ehr}{EHR}{Empty Hyper-Rectangle}
\newacronym{esha}{ESHA}{Environmental Safety Hazard Analysis}

\newacronym{gan}{GAN}{Generative Adversarial Network}

\newacronym{mel}{MEL}{Minimum Equipment List}
\newacronym{ml}{ML}{Machine Learning}

\newacronym{nan}{NaN}{Not a Number}

\newacronym{pca}{PCA}{Principal Component Analysis}

\newacronym{rl}{RL}{Reinforcement Learning}

\newacronym{seu}{SEU}{Single Event Upset}
\newacronym{sqep}{SQEP}{Suitably Qualified and Experienced Personnel}
\newacronym{svm}{SVM}{Support Vector Machine}

\newacronym{wcet}{WCET}{Worst-Case Execution Time}

\begin{document}
\title{Assuring the Machine Learning Lifecycle: Desiderata, Methods, and Challenges}
\author{Rob Ashmore}
\affiliation{%
	\institution{Defence Science and Technology Laboratory, UK}
}
\email{rdashmore@dstl.gov.uk}
\authornote{Authors contributed equally to the paper}
\authornote{Research carried out while the author was a Visiting Fellow of the Assuring Autonomy International Programme.}
\author{Radu Calinescu}
\affiliation{%
	\institution{University of York and Assuring Autonomy International Programme, UK}
}
\email{radu.calinescu@york.ac.uk}
 \authornotemark[1]
\author{Colin Paterson}
\affiliation{%
	\institution{University of York and Assuring Autonomy International Programme, UK}
}
\email{colin.paterson@york.ac.uk}
 \authornotemark[1]

\begin{abstract}
	Machine learning has evolved into an enabling technology for a wide range of highly successful applications. The potential for this success to continue and accelerate has placed machine learning (ML) at the top of research, economic and political agendas. Such unprecedented interest is fuelled by a vision of ML applicability extending to healthcare, transportation, defence and other domains of great societal importance. Achieving this vision requires the use of ML in safety-critical applications that demand levels of assurance beyond those needed for current ML applications. Our paper provides a comprehensive survey of the state-of-the-art in the \emph{assurance of ML}, i.e.\ in the generation of evidence that ML is sufficiently safe for its intended use. 
	The survey covers the methods capable of providing such evidence at different stages of the \emph{machine learning lifecycle}, i.e.\ of the complex, iterative process that starts with the collection of the 	data used to train an ML component for a system, and ends with the deployment of that component within the system. The paper begins with a systematic presentation of the ML lifecycle and its stages. We then define assurance desiderata for each stage, review existing methods that contribute to achieving these desiderata, and identify open challenges that require further research. 
\end{abstract}

\maketitle
\section{Introduction}

The recent success of machine learning (ML) has taken the world by storm. While far from delivering the human-like intelligence postulated by Artificial Intelligence pioneers \cite{Darwiche:2018:HIA:3281635.3271625}, ML techniques such as deep learning have remarkable applications. The use of these techniques in products ranging from smart phones \cite{Anguita2012, Reyes-Ortiz-2016} and household appliances \cite{Kabir-2015} to recommender systems \cite{Ricci-2015, Cheng-2016} and automated translation services \cite{wu-2016} has become commonplace. There is a widespread belief that this is just the beginning of an ML-enabled technological revolution \cite{makridakis2017forthcoming, forsting2017machine}. Stakeholders as diverse as researchers, industrialists, policy makers and the general public envisage that ML will soon be at the core of novel applications and services used in healthcare, transportation, defence and other key areas of economy and society \cite{ komorowski2018artificial,  maurer2016autonomous,  arjomandi2006classification, iskandar2017terrorism,  deng2017deep}. 

Achieving this vision requires a step change in the level of assurance provided 
for ML. The occasional out-of-focus photo taken by an ML-enabled smart camera can easily be deleted, and the selection of an odd word by an automated translation service is barely noticed. Although increasingly rare, similar ML errors would be unacceptable in medical diagnosis applications or self-driving cars. For such safety-critical systems, ML errors can lead to failures that cannot be reverted or ignored, and ultimately cause harm to their users or operators. Therefore, the use of ML to synthesise components of safety-critical systems must be assured by evidence that these \emph{ML components} are fit for purpose \emph{and} adequately integrated into their systems. This evidence must be sufficiently thorough to enable the creation of compelling \emph{assurance cases}~\cite{bloomfield2010safety,SACM-2018} that explain why the systems can be trusted for their intended applications.

Our paper represents the first survey of the methods available for obtaining this evidence for ML components. As with any engineering artefact, assurance can only be provided by understanding the complex, iterative  process employed to produce and use ML components, i.e., the \emph{machine learning lifecycle}. We therefore start by defining this lifecycle, which consists of four stages. The first stage, \emph{Data Management}, focuses on obtaining the data sets required for the training and for the verification of the ML components. This stage includes activities ranging from data collection to data preprocessing \cite{Kotsiantis-etal-2007} (e.g., labelling) and augmentation \cite{ros2016synthia}. The second stage, \emph{Model Learning}, comprises the activities associated with synthesis of the ML component starting from the training data set. The actual machine learning happens in this stage, which also includes activities such as selection of the ML algorithm and hyperparameters \cite{bergstra2012random,thornton2013auto}. The third stage, \emph{Model Verification}, is responsible for providing evidence to demonstrate that the synthesised ML component complies with its requirements. Often treated lightly for ML components used in non-critical applications, this stage is essential for the ML components of safety-critical systems. Finally, the last stage of the ML lifecycle is \emph{Model Deployment}. This stage focuses on the integration and operation of the ML component within a fully-fledged system. 

To ensure a systematic coverage of ML assurance methods, we structure our survey based on the assurance considerations that apply at the four stages of the ML lifecycle. For each stage, we identify the assurance-related \emph{desiderata} (i.e.\ the key assurance requirements, derived from the body of research covered in our survey) for the artefacts produced by that stage. We then present the methods available for achieving these desiderata, with their assumptions, advantages and limitations. This represents an analysis of over two decades of sustained research on ML methods for data management, model learning, verification and deployment. Finally, we determine the open challenges that must be addressed through further research in order to fully satisfy the stage desiderata and to enable the use of ML components in safety-critical systems. 

Our survey and the machine learning lifecycle underpinning its organisation are relevant to a broad range of ML types, including supervised, unsupervised and reinforcement learning. 
Necessarily, some of the methods presented in the survey are only applicable to specific types of ML; we clearly indicate where this is the case. As such, the survey supports a broad range of ML stakeholders, ranging from practitioners developing convolutional neural networks for the classification of road signs in self-driving cars, to researchers devising new ensemble learning techniques for safety-critical applications, and to regulators managing the introduction of systems that use ML components into everyday use. 

The rest of the paper is structured as follows. In Section~\ref{sect:lifecycle}, we present the machine learning lifecycle, describing the activities encountered within each of its stages and introducing ML terminology used throughout the paper. In Section~\ref{sect:related}, we overview the few existing surveys that discuss verification, safety or assurance aspects of machine learning. As we explain in their analysis, each of these surveys focuses on a specific ML lifecycle stage or subset of activities and/or addresses only a narrow aspect of ML assurance. The ML assurance desiderata, methods and open challenges for the four stages of the ML lifecycle are then detailed in Sections~\ref{sect:datamanagement} to~\ref{sect:modeldeployment}. Together, these sections provide a comprehensive set of guidelines for the developers of safety-critical systems with ML components, and inform researchers about areas where additional ML assurance methods are needed. We conclude the paper with a brief summary in Section~\ref{sect:conclusion}.
%!TEX root = ../main.tex

\section{The Machine Learning Lifecycle \label{sect:lifecycle}}

Machine learning represents the automated extraction of \emph{models} (or \emph{patterns}) from data \cite{Bishop2006,goodfellow2016deep,Murphy:2012:MLP:2380985}. In this paper we are concerned with the use of such ML models in safety-critical systems, e.g., to enable these systems to understand the environment they operate in, and to decide their response to changes in the environment. 
Assuring this use of ML models requires an in-depth understanding of the \emph{machine learning lifecycle}, i.e., of the process used for their development and integration into a fully-fledged system. 
Like traditional system development, this process is underpinned by a set of system-level requirements, from which the requirements and operating constraints for the ML models are derived. As an example, the requirements for a ML model for the classification of British road signs can be derived from the high-level requirements for a self-driving car intended to be used in the UK. However, unlike traditional development processes, the development of ML models involves the acquisition of data sets, and \emph{experimentation}~\cite{mitchell1997,zaharia2018accelerating}, i.e., the manipulation of these data sets and the use of ML \emph{training} techniques to produce models of the data that optimise error functions derived from requirements. This experimentation yields a processing pipeline capable of taking data as input and of producing ML models which, when integrated into the system and applied to data unseen during training, achieve their requirements in the deployed context.

\begin{figure}
	\includegraphics[width=\textwidth]{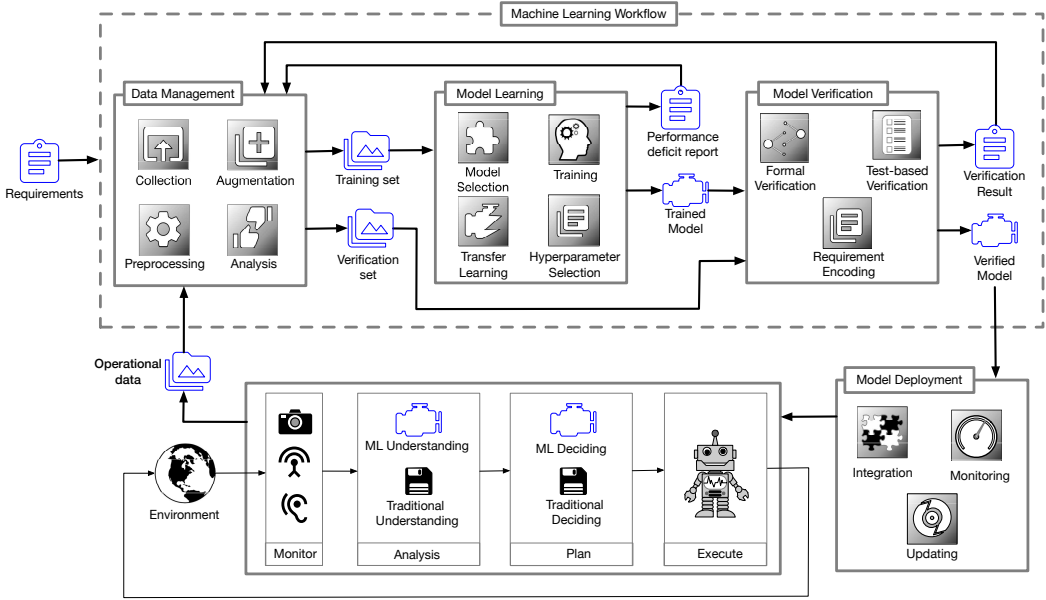}
	
	\vspace*{-1.5mm}
	\caption{The machine learning lifecycle}
	\label{fig:MLLC}
	
	\vspace*{-3mm}
\end{figure}

As shown in Figure~\ref{fig:MLLC}, the machine learning lifecycle consists of four stages. The first three stages---\emph{Data Management}, \emph{Model Learning}, and \emph{Model Verification}---comprise the activities by which machine-learnt models are produced. Accordingly, we use the term \emph{machine learning workflow} to refer to these stages taken together. The fourth stage, \emph{Model Deployment}, comprises the activities concerned with the deployment of ML models within an operational system,  alongside components obtained using traditional software and system engineering methods. We provide brief descriptions of each of these stages below.

Data is at the core of any application of machine learning. As such, the ML lifecycle starts with a Data Management stage. This stage is responsible for the acquisition of the data underpinning the synthesis of machine learnt models that can then be used ``to predict future data, or to perform other kinds of decision making under uncertainty''~\cite{Murphy:2012:MLP:2380985}. This stage comprises four key activities, and produces the \emph{training data set} and \emph{verification data set} used for the training and verification of the ML models in later stages of the ML lifecycle, respectively. The first data management activity, \emph{collection} \cite{wagstaff2012machine, geron2017hands}, is concerned with gathering \emph{data samples} through observing and measuring the real-world (or a representation of the real-world) system, process or phenomenon for which an ML model needs to be built. When data samples are unavailable for certain scenarios, or their collection would be too costly, time consuming or dangerous, \emph{augmentation} methods \cite{wong2016understanding,ros2016synthia} are used to add further data samples to the collected data sets. Additionally, the data collected from multiple sources may be heterogeneous in nature, and therefore \emph{preprocessing} \cite{kotsiantis2006data,zhang2003data} may be required to produce consistent data sets for training and verification purposes. Preprocessing may also seek to reduce the complexity of collected data or to engineer features to aid in training \cite{heaton2016empirical,khurana2018feature}. 
Furthermore, preprocessing may be required to label the data samples when they are used in supervised ML tasks \cite{geron2017hands,goodfellow2016deep, Murphy:2012:MLP:2380985}.
The need for additional data collection, augmentation and preprocessing is established through the \emph{analysis} of the data \cite{r-bloggers}.

In the Model Learning stage of the machine learning lifecycle, the ML engineer typically starts by selecting the type of model to be produced. This \emph{model selection} is undertaken with reference to the problem type (e.g., classification or regression), the volume and structure of the training data \cite{sci-kit-taxonomy,azure-taxonomy}, and often in light of personal experience. A \emph{loss function} is then constructed as a measure of training error. The aim of the \emph{training} activity is to produce an ML model that minimises this error. This requires the development of a suitable data use strategy, so as to determine how much of the training data set should be held for model validation,\footnote{Model validation represents the frequent evaluation of the ML model during training, and is carried out by the development team in order to calibrate the training algorithm. This differs essentially from what validation means in software engineering (i.e., an independent assessment performed to establish whether a system satisfies the needs of its intended users).} and whether all the other data samples should be used together for training or ``minibatch methods'' that use subsets of data samples over successive training cycles should be employed~\cite{goodfellow2016deep}. 
The ML engineer is also responsible for \emph{hyperparameter selection}, i.e., for the choosing the parameters of the training algorithm. Hyperparameters
control key ML model characteristics such as overfitting, underfitting and model complexity.
Finally, when models or partial models that have proved successful within a related context are available, \emph{transfer learning} enables their integration 
within the new model architecture or 
their use as a starting point for training \cite{sukhija2018supervised,ramon2007transfer,oquab2014learning}. When the resulting ML model achieves satisfactory levels of performance, the next stage of the ML worklow can commence. Otherwise, the process needs to return to the Data Management stage, where additional data are collected, augmented, preprocessed and analysed in order to improve the training further.

The third stage of the ML lifecycle is Model Verification. 
The central challenge of this stage is to ensure that the trained model performs well on new, previously unseen inputs (this is known as generalization)~\cite{goodfellow2016deep,Murphy:2012:MLP:2380985,geron2017hands}. As such, the stage comprises activities that provide evidence of the model's ability to generalise to data not seen during the model learning stage. A \emph{test-based verification} activity assesses the performance of the learnt model against the verification data set that the Data Management stage has produced independently from the training data set. 
This data set will have commonalities with the training data, but it may also include elements that have been deliberately chosen to demonstrate a verification aim, which it would be inappropriate to include in the training data. When the data samples from this set are presented to the model, a \emph{generalization error} is computed \cite{niyogi1996relationship,srivastava2014dropout}. If this error violates performance criteria established by a \emph{requirement encoding} activity, then the process needs to return to either the Data Management stage or the Model Learning stage of the ML lifecycle. Additionally, a \emph{formal verification} activity may be used to verify whether the model complies with a set of formal properties that encode key requirements for the ML component.
Formal verification methods such as model checking and mathematical proof 
allow for these properties to be rigorously established before the ML model is deemed suitable for integration into the safety-critical system. As for failed testing-based verification, further Data Management and/or Model Learning activities are necessary when these properties do not hold. The precise activities required from these earlier stages of the ML workflow are determined by the \emph{verification result}, which summarises the outcome of all verification activities. 

Assuming that the verification result contains all the required assurance evidence, a system that uses the now \emph{verified model} is assembled in the Model Deployment stage of the ML lifecycle. This stage comprises activities concerned with the \emph{integration} of verified ML model(s) with system components developed and verified using traditional software engineering methods, with the \emph{monitoring} of its operation, and with its \emph{updating} thorough offline maintenance or online learning. The outcome of the Model Deployment stage is a fully-fledged deployed and operating system. 

More often than not, the safety-critical systems envisaged to benefit from the use of ML models are autonomous or self-adaptive systems that require ML components to cope with the dynamic and uncertain nature of their operating environments \cite{maurer2016autonomous,komorowski2018artificial,8008800}. As such, Figure~\ref{fig:MLLC} depicts this type of system as the outcome of the Model Deployment stage. Moreover, the diagram shows two key roles that ML models 
may play within the established \emph{monitor-analyse-plan-execute} (MAPE) control loop \cite{kephart2003vision,iglesia2015mape,maurer2011revealing} of these systems. We end this section with a brief description of the MAPE control loop and of these typical uses of ML models within it.

In its four steps, the MAPE control loop senses the current state of the environment through \emph{monitoring}, derives an understanding of the world through the \emph{analysis} of the sensed data, decides suitable actions through \emph{planning}, and then acts upon these plans through \emph{executing} their actions. Undertaking these actions alters the state of the system and the environment in which it operates.

The monitoring step employs hardware and software components that gather data as a set of samples from the environment during operation. The choice of sensors requires an understanding of the system requirements, the intended operational conditions, and the platform into which they will be deployed. Data gathered from the environment will typically be partial and imperfect due to physical, timing and financial constraints.

The analysis step extracts features from data samples as encoded domain-specific knowledge. This can be achieved through the use of ML models combined with traditional software components. The features extracted may be numerical (e.g., blood sugar level in a healthcare system), ordinal (e.g., position in queue in a traffic management system) or categorical (e.g., an element from the set $\{\mathsf{car}, \mathsf{bike}, \mathsf{bus}\}$ in a self-driving car). The features extracted through analysing the data sets obtained during monitoring underpin the understanding of the current state of the environment and that of the system itself.

The planning (or decision) step can employ a combination of ML models and traditional reasoning engines in order to select a course of action to be undertaken. The action(s) selected will aim to fulfil the system requirements subject to any defined constraints. The action set available is dictated by the capabilities of the system, and is restricted by operating conditions and constraints defined in the requirements specification.

Finally, in the execution step, the system enacts the selected actions through software and hardware effectors and, in doing so, changes the environment within which it is operating. The dynamic and temporal nature of the system and the environment requires the MAPE control loop to be invoked continuously until a set of system-level objectives has been achieved or a stopping criterion was reached. 

New data samples gathered during operation can be exploited by the Data Management activities and, where appropriate, new models may be learnt and deployed within the system. This deployment of new ML models can be carried out either as an offline maintenance activity, or through the online updating of the operating system.
%!TEX root = ../main.tex

\section{Related Surveys \label{sect:related}}

The large and rapidly growing body of ML research is summarised by a plethora of surveys. The vast majority of these surveys narrowly focus on a particular type of ML, and do not consider assurance explicitly. Examples range from surveys on deep learning \cite{deng2014tutorial,liu2017survey,pouyanfar2018survey} and reinforcement learning \cite{kaelbling1996reinforcement,kober2013reinforcement} to surveys on transfer learning \cite{cook2013transfer,lu2015transfer,weiss2016survey} and ensemble learning \cite{gomes2017survey,krawczyk2017ensemble,mendes2012ensemble}. These surveys provide valuable insights into the applicability, effectiveness and trade-offs of the Data Management and Model Learning methods available for the type of ML they cover. However, they do not cover the assurance of the ML models obtained by using these methods.

A smaller number of surveys consider a wider range of ML types but focus on a single stage, or on an activity from a stage, of the ML lifecycle, again without addressing assurance explicitly. Examples include surveys on the Data Management stage  \cite{roh2018survey,Polyzotis:2018:DLC:3299887.3299891}, and surveys on feature selection \cite{chandrashekar2014survey,elavarasan2015survey,khalid2014survey} and dimensionality reduction \cite{camastra2003data,cunningham2015linear,wang2015survey} within the Model Learning stage, respectively. These surveys identify effective methods for the specific ML lifecycle stage or activity they concentrate on, but are complementary to the survey presented in our paper.

A key assurance-related property of ML models, interpretability, has been the focus of intense research in recent years, and several surveys of the methods devised by this research are now available ~\cite{zhang2018visual,dovsilovic2018explainable,adadi2018peeking}. Unlike our paper, these surveys do not cover other assurance-related desiderata of the models produced by the Model Learning stage (cf.~Section~\ref{sect:modellearning}), nor the key properties of the artefacts devised by the other stages of the ML lifecycle.

Only a few surveys published recently address the generation of assurance evidence for machine learning \cite{xiang2018verification} and safety in the context of machine learning \cite{garcia2015comprehensive,salay2018using,corr/abs-1812-08342}, respectively. We discuss each of these surveys and how it relates to our paper in turn. 

First, the survey in \cite{xiang2018verification} covers techniques and tools for the verification of neural networks (with a focus on formal verification), and approaches to implementing the intelligent control of autonomous systems using neural networks. In addition to focusing on a specific class of verification techniques and a particular type of ML, this survey does not cover assurance-relevant methods for the Data Management and Model Learning stages of the ML lifecycle, and only briefly refers to the safety concerns of integrating ML components into autonomous systems. In contrast, our paper covers all these ML assurance aspects systematically.

The survey on safe reinforcement learning (RL) by Garc\'{i}a and Fern\'{a}ndez \cite{garcia2015comprehensive} overviews methods for devising RL policies that reduce a risk metric or that satisfy predefined safety-related constraints. Both methods that work by modifying the RL optimisation criterion and methods that adjust the RL exploration are considered. However, unlike our paper, the survey in \cite{garcia2015comprehensive} only covers the Model Learning stage of the ML lifecycle, and only for reinforcement learning.

Huang \emph{et al}'s recent survey \cite{corr/abs-1812-08342} provides an extensive coverage of formal verification and testing methods for deep neural networks. The survey also overviews adversarial attacks on deep neural networks, methods for defence against these attacks, and interpretability methods for deep learning. This specialised survey provides very useful guidelines on the methods that can be used in the verification stage of the ML lifecycle for deep neural networks, but does not look at the remaining stages of the lifecycle and at other types of ML models like our paper.

Finally, while not a survey per se, Salay and Czarnecki's methodology for the assurance of ML safety in automotive software \cite{salay2018using} discusses existing methods that could support supervised-learning assurance within multiple activities from the ML lifecycle. Compared to our survey, \cite{salay2018using} typically mentions a single assurance-supporting method for each such activity, does not systematically identify the stages and activities of the ML lifecycle, and is only concerned with supervised learning for a specific application domain.
%!TEX root = ../main.tex

\section{Data Management} \label{sect:datamanagement}

Fundamentally, all ML approaches start with data. These data describe the desired relationship between the ML model inputs and outputs, the latter of which may be implicit for unsupervised approaches. Equivalently, these data encode the requirements we wish to be embodied in our ML model. Consequently, any assurance argument needs to explicitly consider data.

\subsection{Inputs and Outputs}

The key input artefact to the Data Management stage is the set of requirements that the model is required to satisfy. These may be informed by verification artefacts produced by earlier iterations of the ML lifecycle. The key output artefacts from this stage are data sets: there is a combined data set that is used by the development team for training and validating the model; there is also a separate verification data set, which can be used by an independent verification team.

\subsection{Activities}

\subsubsection{Collection}
This  activity is concerned with collecting data from an originating source. These data may be subsequently enhanced by other activities within the Data Management stage. New data may be collected, or a pre-existing data set may be re-used (or extended). Data may be obtained from a controlled process, or they may arise from observations of an uncontrolled process: this process may occur in the real world, or it may occur in a synthetic environment.

\subsubsection{Preprocessing}
For the purposes of this paper we assume that preprocessing is a one-to-one mapping: it adjusts each collected (raw) sample in an appropriate manner. It is often concerned with standardising the data in some way, e.g., ensuring all images are of the same size \cite{lecun1998gradient}. Manual addition of labels to collected samples is another form of preprocessing.

\subsubsection{Augmentation\label{sec:augmentation}}
Augmentation increases the number of samples in a data set. Typically, new samples are derived from existing samples, so augmentation is, generally, a one-to-many mapping. Augmentation is often used due to the difficulty of collecting observational data (e.g., for reasons of cost or ethics \cite{ros2016synthia}). Augmentation can also be used to help instil certain properties in the trained model, e.g., robustness to adversarial examples \cite{goodfellow2014explaining}.

\subsubsection{Analysis}
Analysis may be required to guide aspects of collection and augmentation (e.g., to ensure there is an appropriate class balance within the data set). Exploratory analysis is also needed to provide assurance that Data Management artefacts exhibit the  desiderata below.

\subsection{Desiderata} \label{subsect:dm_desiderata}

From an assurance perspective, the data sets produced during the Data Management stage should exhibit the following key properties:

\begin{enumerate}
  \item \textsf{Relevant}---This property considers the intersection between the data set and the desired behaviour in the intended operational domain. For example, a data set that only included German road signs would not be \textsf{Relevant} for a system intended to operate on UK roads.  
  \item \textsf{Complete}---This property considers the way samples are distributed across the input domain and subspaces of it. In particular, it considers whether suitable distributions and combinations of features are present. For example, an image data set that displayed an inappropriate correlation between image background and type of animal would not be complete \cite{ribeiro2016should}.
  \item \textsf{Balanced}---This property considers the distribution of features that are included in the data set. For classification problems, a key consideration is the balance between the number of samples in each class \cite{haixiang2017learning}. This property takes an internal perspective; it focuses on the data set as an abstract entity. In contrast, the \textsf{Complete} property takes an external perspective; it considers the data set within the intended operational domain.
  \item \textsf{Accurate}---This property considers how measurement (and measurement-like) issues can affect the way that samples reflect the intended operational domain. It covers aspects like sensor accuracy and labelling errors \cite{brodley1999identifying}. The correctness of data collection and preprocessing software is also relevant to this property, as is configuration management.
\end{enumerate}

Conceptually, since it relates to real-world behaviour, the \textsf{Relevant} desideratum is concerned with validation. The other three desiderata are concerned with aspects of verification.

\subsection{Methods}

This section considers each of the four desiderata in turn. Methods that can be applied during each Data Management activity, in order to help achieve the desired key property, are discussed.

\subsubsection{Relevant} \label{subsubsect:relevant}

By definition, a data set collected during the operational use of the planned system will be relevant.  However, this is unlikely to be a practical way of obtaining all required data.

If the approach adopted for data collection involves re-use of a pre-existing data set, then it needs to be acquired from an appropriate source. Malicious entries in the data set can introduce a backdoor, which causes the model to behave in an attacker-defined way on specific inputs (or small classes of inputs) \cite{chen2017backdoor}. In general, detection of hidden backdoors is an open challenge (listed as DM01 in Table~\ref{tab:dm_chall} at the end of Section~\ref{sect:datamanagement}). It follows that pre-existing data sets should be obtained from trustworthy sources via  means that provide strong guarantees on integrity during transit.

If the data samples are being collected from controlled trials, then we would require an appropriate experimental plan that justifies the choice of feature values (inputs) included in the trial. If the trial involves real-world observations then traditional experimental design techniques will be appropriate \cite{kirk2007experimental}. Conversely, if the trial is conducted entirely in a simulated environment then techniques for the design and analysis of computer experiments will be beneficial \cite{sacks1989design}. 

If the data set contains synthetic samples (either from collection or as a result of augmentation) then we would expect evidence that the synthesis process is appropriately representative of the real-world. Often, synthesis involves some form of simulation, which ought to be suitably verified and validated \cite{sargent2009verification}, as there are examples of ML-based approaches affected by simulation bugs \cite{chrabaszcz2018back}. Demonstrating that synthetic data is appropriate to the real-world intended operational domain, rather than a particular idiosyncrasy of a simulation, is an open challenge (DM02 in Table~\ref{tab:dm_chall}).

A data set can be made irrelevant by data leakage. This occurs when the training data includes information that will be unavailable to the system within which the ML model will be used \cite{kaufman2012leakage}. One way of reducing the likelihood of leakage is to only include in the training data features that can ``legitimately'' be used to infer the required output. For example, patient identifiers are unlikely to be legitimate features for any medical diagnosis system, but may have distinctive values for patients already diagnosed with the condition that the ML model is meant to identify \cite{rosset2010medical}. Exploratory data analysis (EDA) \cite{tukey1977exploratory} can help identify potential sources of leakage: a surprising degree of correlation between a feature and an output may be indicative of leakage. That said, detecting and correcting for data leakage is an open challenge (DM03 in Table~\ref{tab:dm_chall}).

Although it appears counter-intuitive, augmenting a data set by including samples that are highly unlikely to be observed during operational use can increase relevance.  For classification problems, adversarial inputs \cite{szegedy2013intriguing} are specially crafted inputs that a human would classify correctly but are confidently mis-classified by a trained model. Including adversarial inputs \emph{with the correct class} in the training data \cite{papernot2017practical} can help reduce mis-classification and hence increase relevance. Introducing an artificial \emph{unknown}, or ``dustbin'', class and augmenting the data with suitably placed samples attributed to this class \cite{abbasi2018controlling} can also help.

Finally, \emph{unwanted bias} (i.e., systematic error ultimately leading to unfair advantage for a privileged class of system users) can significantly impact the relevance of a data set. This can be addressed using pre-processing techniques that remove the predictability of data features such as ethnicity, age or gender \cite{feldman2015certifying} or augmentation techniques that involve data relabelling/reweighing/re\-sampling \cite{kamiran2012data}. It can also be addressed during the Model Learning and Model Deployment stages. An industry-ready toolkit that implements a range of methods for addressing unwanted bias is available \cite{bellamy2018ai}.

\subsubsection{Complete} \label{subsubsect:complete}

Recall that this property is about how the data set is distributed across the input domain. For the purposes of our discussion, we define four different spaces related to that domain:

\begin{enumerate}
  \item The \emph{input domain} space, $\mathcal{I}$, which is the set of inputs that the model can accept. Equivalently, this set is defined by the input parameters of the software implementation that instantiates the model. For example, a model that has been trained on grey-scale images may have an input space of $256 \times 256 \times \texttt{UINT8}$; that is, a 256 by 256 square of unsigned 8-bit integers.

  \item The \emph{operational domain} space, $\mathcal{O} \subset \mathcal{I}$, which is the set of inputs that the model may be expected to receive when used within the intended operational domain. In some cases, it may be helpful to split $\mathcal{O}$ into two subsets: inputs that can be (or have been) acquired through collection and inputs that can only be (or have been) generated by augmentation.

  \item The \emph{failure domain} space, $\mathcal{F} \subset \mathcal{I}$, which is the set of inputs the model may receive if there are failures elsewhere in the system. The distinction between $\mathcal{F}$ and $\mathcal{O}$ is best conveyed by noting that $\mathcal{F}$ covers system states, whilst $\mathcal{O}$ covers environmental effects: a cracked camera lens should be covered in $\mathcal{F}$; a fly on the lens should be covered in $\mathcal{O}$. 

  \item The \emph{adversarial domain} space, $\mathcal{A} \subset \mathcal{I}$, which is the set of inputs the model may receive if it is being attacked by an adversary. This includes adversarial examples, where small changes to an input cause misclassification \cite{szegedy2013intriguing}, as well as more general cyber-related attacks.
\end{enumerate}

The consideration of whether a data set is complete with regards to the input domain can be informed by simple statistical analysis and EDA \cite{tukey1977exploratory}, supported by discussions with experts from the intended operational domain. Simple plots showing the marginal distribution of each feature can be surprisingly informative. Similarly, the ratio of sampling density between densely sampled and sparsely sampled regions is informative \cite{bishnu2015gapratio} as is, for classification problems, identifying regions that only contain a single class \cite{ashmore2018boxing}. Identifying any large empty hyper-rectangles (EHRs) \cite{lemley2016holes}, which are large regions without any samples, is also important. If an operational input comes from the central portion of a large EHR then, generally speaking, it is appropriate for the system to know the model is working from an area for which no training data were provided.

Shortfalls in completeness across the input domain can be addressed via collection or augmentation. Since a shortfall will relate to a lack of samples in a specific part of the input domain, further collection is most appropriate in the case of controlled trials.

Understanding completeness from the perspective of the \emph{operational domain} space is challenging. Typically, we would expect the input space $\mathcal{I}$ to be high-dimensional, with $\mathcal{O}$ being a much lower-dimensional manifold within that space \cite{saul2003think}. Insights into the scope of $\mathcal{O}$ can be obtained by requirements decomposition. The notion of situation coverage \cite{alexander2015situation} generalises these considerations. 

If an increased coverage of $\mathcal{O}$ is needed, then this could be achieved via the use of a generative adversarial network (GAN)\footnote{A GAN is a network specifically designed to provide inputs for another network. A classification network tries to learn the boundary between classes, whereas a GAN tries to learn the distribution of individual classes.} \cite{antoniou2017data} that has been trained to model the distributions of each class.

Although the preceding paragraphs have surveyed multiple methods, understanding completeness across the operational domain remains an open challenge (DM04 in Table~\ref{tab:dm_chall}). 

Completeness across the $\mathcal{F}$ space can be understood by systematically examining the system architecture to identify failures that could affect the model's input. Note that the system architecture may protect the model from the effects of some failures: for example, the system may not present the model with images from a camera that has failed its built-in test. In some cases it may be possible to collect samples that relate to particular failures. However, for reasons of cost, practicality and safety, augmentation is likely to be needed to achieve suitable completeness of this space \cite{alhaija2018augmented}. Finding verifiable ways of achieving this augmentation is an open challenge (DM05 in Table~\ref{tab:dm_chall}).

Understanding completeness across the adversarial domain $\mathcal{A}$ involves checking: the model's susceptibility to known ways of generating adversarial examples \cite{szegedy2013intriguing, goodfellow2014explaining,moosavi2017universal,yuan2018song}; and its behaviour when presented with inputs crafted to achieve some other form of behaviour, for example, a not-a-number (NaN) error. Whilst they are useful, both of these methods are subject to the ``unknown unknowns'' problem. More generally, demonstrating completeness across the adversarial domain is an open challenge (DM06 in Table~\ref{tab:dm_chall}).

\subsubsection{Balanced} \label{subsubsect:balanced}

This property is concerned with the distribution of the data set, viewed
from an internal perspective. Initially, it is easiest to think about balance solely from the perspective of supervised classification, where a key consideration is the number of samples in each class. If this is unbalanced then simple measures of performance (e.g., classifier accuracy) may be insufficient \cite{lopez2013insight}. 

As it only involves counting the number of samples in each class, detecting class imbalance is straightforward. Its effects can be countered in several ways; for example, in the Model Learning and Model Verification stages, performance measures can be class-specific, or weighted to account for class imbalance \cite{haixiang2017learning}. Importance weighting can, however, be ineffective for deep networks trained for many epochs \cite{LiptonImportanceWeighting}. Alternatively, or additionally, in the Data Management stage augmentation can be used to correct (or reduce) the class imbalance, either by oversampling the minority class, or by undersampling the majority class, or using a combination of these approaches\footnote{Note that the last two approaches involve removing samples (corresponding to the majority class) from the data set; this differs from the normal view whereby augmentation increases the number of samples in the data set.} \cite{lopez2013insight}. If data are being collected from a controlled trial then another approach to addressing class imbalance is to perform additional collection, targeted towards the minority class.

Class imbalance can be viewed as being a special case of rarity \cite{weiss2004mining}. Another way rarity can manifest is through small disjuncts, which are small, isolated regions of the input domain that contain a single class. Analysis of single-class regions \cite{ashmore2018boxing} can inform the search for small disjuncts, as can EDA and expertise in the intended operational domain. Nevertheless, finding small disjuncts remains an open challenge (DM07 in Table~\ref{tab:dm_chall}). 

Class imbalance can also be viewed as a special case of a phenomenon that applies to all ML approaches: feature imbalance. Suppose we wish to create a model that applies to people of all ages. If almost all of our data relate to people between the ages of 20 and 40, we have an imbalance in this feature. Situations like this are not atypical when data is collected from volunteers. Detecting such imbalances is straightforward; understanding their influence on model behaviour and correcting for them are both open challenges (DM08 and DM09 in Table~\ref{tab:dm_chall}).

\subsubsection{Accurate} \label{subsubsect:accurate}
Recall that this property is concerned with measurement (and measurement-like) issues. If sensors are used to record information as part of data collection then both sensor precision and accuracy need to be considered. If either of these is high, there may be benefit in augmenting the collected data with samples drawn from a distribution that reflects precision or accuracy errors.

The actual value of a feature is often unambiguously defined \cite{smyth1996bounds}. However, in some cases this may not be possible: for example, is a person walking astride a bicycle a pedestrian or a cyclist? Consequently, labelling discrepancies are likely, especially when labels are generated by humans. Preventing, detecting and resolving these discrepancies is an open challenge (DM10 in Table~\ref{tab:dm_chall}).

The data collection process should generally be documented in a way 
%Generally speaking, we would expect there to be a documented process for data collection 
that accounts for potential weaknesses in the approach.  If the process uses manual recording of information, we would expect steps to be taken to ensure attention does not waver and records are accurate. Conversely, if the data collection process uses logging software, confidence that this software is behaving as expected should be obtained, e.g. using traditional approaches for software quality \cite{ISO26262,RTCA2011178C}. 
	
This notion of correct behaviour applies across all software used in the Data Management stage (and to all software used in the ML lifecycle). Data collection software may be relatively simple, merely consisting of an automatic recording of sensed values. Alternatively, it may be very complex, involving a highly-realistic simulation of the intended operational domain. The amount of evidence needed to demonstrate correct behaviour is related to the complexity of the software. Providing sufficient evidence for a complex simulation is an open challenge (DM11 in Table~\ref{tab:dm_chall}).

Given their importance, data sets should be protected against unintentional and unauthorised changes. Methods used in traditional software development (e.g., \cite{ISO26262,RTCA2011178C}) may be appropriate for this task, but they may be challenged by the large volume and by the non-textual nature of many of the data sets used in ML.

\subsection{Summary and Open Challenges}

Table~\ref{tab:dm_astech} summarises the assurance methods that can be applied during the Data Management stage. For ease of reference, the methods are presented in the order they were introduced in the preceding discussion.  Methods are also matched to activities and desiderata.

Table~\ref{tab:dm_astech} shows that there are relatively few methods associated with the preprocessing activity. This may be because preprocessing is, inevitably, problem-specific. Likewise, there are few methods associated with the \textsf{Accurate} desideratum. This may reflect the widespread use of commonly available data sets (e.g., ImageNet \cite{deng2009imagenet} and MNIST) within the research literature, which de-emphasises issues associated with data collection and curation, like \textsf{Accuracy}.

\begin{table}
	\centering
	\caption{Assurance methods for the Data Management stage}
	\begin{footnotesize}
	
	\vspace*{-2mm}
	\def\tabcolsep{1.4pt}
	\sffamily
	\begin{tabular}{L{4.25cm}cccccccc} 
	\toprule
        & \multicolumn{4}{c}{\textbf{Associated activities$^\dagger$}} & \multicolumn{4}{c}{\textbf{Supported desiderata$^\ddagger$}}\\ \cmidrule(l{1pt}r{2pt}){2-5}\cmidrule(l{2pt}r{1pt}){6-9}
		\textbf{Method} & \hspace*{-0.3mm}Collection & Preprocess. & Augment. & Analysis & Relevant & Complete & Balanced & Accurate \\ \midrule
		\\[-1.5em]
   	    \rowcolor{gray!25}
		Use trusted data sources, with data-transit integrity guarantees & \ding{52} & & & & \ding{72} & & & \\ 
		Experimental design \cite{kirk2007experimental}, \cite{sacks1989design} & \ding{52} & & \ding{52} & & \ding{72} & \ding{72} & \ding{73} & \\ 
   	    \rowcolor{gray!25}
		Simulation verification and validation \cite{sargent2009verification} & & & \ding{52} & & \ding{72} & \ding{73} & \ding{73} & \\ 
		Exploratory data analysis \cite{tukey1977exploratory} & & & & \ding{52} & & \ding{72} & \ding{72} & \\ 
   	    \rowcolor{gray!25}
		Use adversarial examples \cite{papernot2017practical} & & & \ding{52} & & \ding{73} & \ding{72} & & \\ 
		Include a ``dustbin'' class \cite{abbasi2018controlling} & & & \ding{52} & & \ding{73} & \ding{72} & & \\ 
   	    \rowcolor{gray!25}
		Remove unwanted bias \cite{bellamy2018ai} & & \ding{52} & \ding{52} & & \ding{72} & & \ding{73} & \\
		Compare sampling density \cite{bishnu2015gapratio} & & & \smltick & \ding{52} & & \ding{72} & \ding{73} & \\ 
   	    \rowcolor{gray!25}
		Identify empty and single-class regions \cite{lemley2016holes}, \cite{ashmore2018boxing} & & & \smltick & \ding{52} & & \ding{72} & \ding{73} & \\ 
		Use situation coverage \cite{alexander2015situation} & & & & \ding{52} & & \ding{72} & & \\ 
   	    \rowcolor{gray!25}
		Examine system failure cases & & & & \ding{52} & & \ding{72} & & \\ 
		Oversampling \hspace*{-0.2mm}\&\hspace*{-0.2mm} undersampling \cite{lopez2013insight} & & & & \ding{52} & & \ding{72} & \ding{72} & \\ 
   	    \rowcolor{gray!25}
		Check for within-class \cite{japkowicz2001concept} and feature imbalance & & & & \ding{52} & & \ding{72} & & \\ 
		Use a GAN \cite{antoniou2017data} & & & \ding{52} & & & \ding{72} & \ding{73} & \\ 
   	    \rowcolor{gray!25}
		Augment data to account for sensor errors & \smltick & & \ding{52} & & \ding{73} & & & \ding{72} \\
		Confirm correct software behaviour \cite{ISO26262}, \cite{RTCA2011178C} & \smltick & \ding{52} & \ding{52} & \smltick & \ding{73} & \ding{72} & \ding{73} & \ding{73} \\ 
   	    \rowcolor{gray!25}
		Use documented processes & \ding{52} & \ding{52} & \ding{52} & \ding{52} & \ding{73} & & & \ding{72} \\ 
		Apply configuration management \cite{ISO26262}, \cite{RTCA2011178C} & \ding{52} & \ding{52} & \ding{52} & \ding{52} & \ding{73} & & & \ding{72} \\[-0.5mm] \bottomrule
		\multicolumn{9}{l}{$^\dagger$\ding{52} = activity that the method is typically used in; \smltick = activity that may use the method}\\
		\multicolumn{9}{l}{$^\ddagger$\ding{72} = desideratum supported by the method; \ding{73} = desideratum partly supported by the method}
	\end{tabular}
	\end{footnotesize}
	\label{tab:dm_astech}
\end{table}

\begin{table}
  \centering
		\caption{Open challenges for the assurance concerns associated with the Data Management (DM) stage\label{tab:DMChallenges}}
		
	\vspace*{-2mm}
	\begin{footnotesize}
	\sffamily
  	\begin{tabular}{L{0.75cm}L{8.5cm}L{2.9cm}}
  	  \toprule
  	  \textbf{ID} & \textbf{Open Challenge} & \textbf{Desideratum (Section)}\\ \midrule
			DM01 & Detecting backdoors in data & \multirow{3}{*}{Relevant (Section \ref{subsubsect:relevant})} \\
			DM02 & Demonstrating synthetic data appropriateness to the operational domain \\
			DM03 & Detecting and correcting for data leakage \\ \midrule
			DM04 & Measuring completeness with respect to the operational domain & \multirow{3}{*}{Complete (Section \ref{subsubsect:complete})} \\
			DM05 & Deriving ways of drawing samples from the failure domain \\
			DM06 & Measuring completeness with respect to the adversarial domain \\ \midrule
			DM07 & Finding small disjuncts, especially for within-class imbalances & \multirow{3}{*}{Balanced (Section \ref{subsubsect:balanced})} \\
			DM08 & Understanding the effect of feature imbalance on model performance \\
			DM09 & Correcting for feature imbalance \\ \midrule
			DM10 & Maintaining consistency across multiple human collectors/preprocessors & \multirow{2}{*}{Accurate (Section \ref{subsubsect:accurate})} \\ 
			DM11 & Verifying the accuracy of a complex simulation \\ \bottomrule
  	\end{tabular}
  	\end{footnotesize}
		\label{tab:dm_chall}
\end{table}

Open challenges associated with the Data Management stage are shown in Table~\ref{tab:dm_chall}. The relevance and nature of these challenges have been established earlier in this section. For ease of reference, each  challenge is matched to the artefact desideratum that it is most closely related to. It is apparent that, with the exception of understanding the effect of feature imbalance on model performance, these open challenges do \emph{not} relate to the core process of learning a model. As such, they emphasise important areas that are insufficiently covered in the research literature. Examples include being able to demonstrate: that the model is sufficiently secure---from a cyber perspective (open challenge DM01); that the data are fit-for-purpose (DM02, DM03); that the data cover operational, failure and adversarial domains (DM04, DM05, DM06); that the data are balanced, across and within classes (DM07, DM08, DM09); that manual data collection has not been compromised (DM10); and that simulations are suitably detailed and representative of the real world (DM11).

%!TEX root = ../main.tex

\section{Model Learning} \label{sect:modellearning}

The Model Learning stage of the ML lifecycle is concerned with creating a model, or algorithm, from the data presented to it. A good model will replicate the desired relationship between inputs and outputs present in the training set, and will satisfy non-functional requirements such as providing an output within a given time and using an acceptable amount of computational resources.

\subsection{Inputs and Outputs}

The key input artefact to this stage is the training data set produced by the Data Management stage. The key output artefacts are a machine-learnt model for verification in the next stage of the ML lifecycle and a performance deficit report used to inform remedial data management activities.

\subsection{Activities}

\subsubsection{Model Selection}

This activity decides the model type, variant and, where applicable, the structure of the model to be produced in the Model Learning stage. Numerous types of ML models are available~\cite{sci-kit-taxonomy,azure-taxonomy}, including multiple types of \emph{classification} models (which identify the category that the input belongs to), \emph{regression} models (which predict a continuous-valued attribute), \emph{clustering} models (which group similar items into sets), and \emph{reinforcement learning} models (which provide an optimal set of actions, i.e.\ a policy, for solving, for instance, a navigation or planning problem). 

\subsubsection{Training}

This activity optimises the performance of the ML model with respect to an objective function that reflects the requirements for the model. To this end, a subset of the training data is used to find internal model parameters (e.g., the weights of a neural network, or the coefficients of a polynomial) that minimise an error metric for the given data set. The remaining data (i.e, the validation set) are then used to assess the ability of the model to generalise. These two steps are typically iterated many times, with the training hyperparameters tuned between iterations so as to further improve the performance of the model. 

\subsubsection{Hyperparameter Selection}

This activity is concerned with selecting the parameters associated with the training activity, i.e., the hyperparameters. Hyperparameters control the effectiveness of the training process, and ultimately the performance of the resulting model~\cite{probst2018tunability}. They are so critical to the success of the ML model that they are often deemed confidential for models used in proprietary systems~\cite{wang2018stealing}. There is no clear consensus on how the hyperparameters should be tuned~\cite{lujan2018design}. Typical options include: initialisation with values offered by ML frameworks; manual configuration based on recommendations from literature or experience; or trial and error~\cite{probst2018tunability}. Alternatively, the tuning of the hyperparameters can itself be seen as a machine learning task~\cite{hutter2015beyond, young2015optimizing}. 

\subsubsection{Transfer Learning}

The training of complex models may require weeks of computation on many GPUs~\cite{gu2017badnets}. As such, there are clear benefits in reusing ML models across multiple domains.  Even when a model cannot be transferred between domains directly, one model may provide a starting point for training a second model, significantly reducing the training time. The activity concerned with reusing models in this way is termed transfer learning~\cite{goodfellow2016deep}.

\subsection{Desiderata}
From an assurance viewpoint, the models generated by the Model Learning stage should exhibit the key properties described below:

\begin{enumerate}
	\item \textsf{Performant}---This property considers quantitative performance metrics applied to the model when deployed within a system. These metrics include traditional ML metrics such as classification accuracy, ROC and mean squared error, as well as metrics that consider the  system and environment into which the models are deployed.  
	\item \textsf{Robust}---This property considers the model's ability to perform well in circumstances where the inputs encountered at run time are different to those present in the training data. Robustness may be considered with respect to environmental uncertainty, e.g. flooded roads, and system-level variability, e.g. sensor failure.
	\item \textsf{Reusable}---This property considers the ability of a model, or of components of a model, to be reused in  systems for which they were not originally intended. For example, a neural network trained for facial recognition in an authentication system may have features which can be reused to identify operator fatigue.
	\item \textsf{Interpretable}---This property considers the extent to which the model can produce artefacts that support the analysis of its output, and thus of any decisions based on it. For example, a decision tree may support the production of a narrative explaining the decision to hand over control to a human operator.  
\end{enumerate}

\subsection{Methods}
This section considers each of the four desiderata in turn. Methods applicable during each of the Model Learning activities, in order to help achieve each of the desired properties, are discussed.

\subsubsection{Performant\label{subsubsect:performant}}

An ML model is performant if it operates as expected according to a measure (or set of measures) that captures relevant characteristics of the model output. Many machine learning problems are phrased in terms of objective functions to be optimized~\cite{wagstaff2012machine}, and measures constructed with respect to these objective functions allow models to be compared.
Such measures have underpinning assumptions and limitations which should be fully understood before they are used to select a model for deployment in a safety-critical system.  

The prediction error of a model  has three components: \emph{irreducible error}, which cannot be eliminated regardless of the algorithm or training methods employed; \emph{bias error}, due to simplifying assumptions intended to make learning the model easier; and \emph{variance error}, an estimate of how much the model output would vary if different data were used in the training process. The aim of training is to minimise the bias and variance errors, and therefore the objective functions reflect these errors. The objective functions may also contain simplifying assumptions to aid optimization, and these assumptions must not be present when assessing model performance~\cite{geron2017hands}.

Performance measures for classifiers, including accuracy, precision, recall (sensitivity) and specificity, are often derived from their confusion matrix~\cite{geron2017hands,Murphy:2012:MLP:2380985,sokolova2009systematic}. Comparing models is not always straightforward, with different models showing superior performance against different measures. Composite metrics~\cite{geron2017hands, sokolova2009systematic} allow for a  trade-off between measures during the training process.
The understanding of evaluation measures has improved over the  past two decades but areas where understanding is lacking still exist~\cite{flach2019performance}. While using a single, scalar measure simplifies the selection of a ``best'' model and is a common practice~\cite{drummond2006cost}, the ease with which such performance measures can be produced has led to over-reporting  of simple metrics without an explicit statement of their relevance to the operating domain. Ensuring that reported measures convey sufficient contextually relevant information remains an open challenge (challenge ML01 from Table~\ref{tab:MLChallenges}).

Aggregated measures cannot evaluate models effectively except in the simplest scenarios, and the operating environment influences the required trade-off between performance metrics. The receiver operator characteristic (ROC) curve~\cite{provost1998case} allows for Pareto-optimal model selection using a trade-off between the true and false positive rates, while  the area under the ROC curve (AUC)~\cite{bradley1997use}   assesses the sensitivity of models to changes in operating conditions. Cost curves~\cite{drummond2006cost} allow weights to be associated with true and false positives to reflect their importance in the operating domain. Where a single classifier cannot provide an acceptable trade-off, models identified using the ROC curve may be combined to produce a classifier with better performance than any single model, under real-world operating conditions~\cite{provost2001robust}. This requires trade-offs to be decided at training time, which is unfeasible for dynamic environments and  multi-objective optimisation problems. Developing methods to defer this decision until run-time is an open challenge (ML02 in Table~\ref{tab:MLChallenges}).
 
Whilst the measures presented thus far give an indication of the performance of the model against data sets, they do not encapsulate the users' trust in a model for a specific, possibly rare, operating point. The intuitive certainty measure (ICM)~\cite{van2018icm} is a mechanism to produce an estimate of how certain an ML model is for a specific output based on errors made in the past. ICM compares current and previous sensor data to assess similarity, using previous outcomes for similar environmental conditions to inform trust measures. Due to the probabilistic nature of machine learning~\cite{Murphy:2012:MLP:2380985}, models may also be evaluated using classical statistical methods. These methods can answer several key questions~\cite{mitchell1997}: (i)~given the observed accuracy of a model, how well is it likely to estimate unseen samples? (ii)~if a model outperforms another for a specific dataset, how likely is it to be more accurate in general? and (iii)~what is the best way to learn a hypothesis from limited data?

%Having considered methods appropriate for assessing the performance of machine\textcolor{blue}{-}learnt models we now consider model training methods used 
Methods are also available to improve the performance of the ML models. Ensemble learning~\cite{sagi2018ensemble} combines multiple models to produce a model whose performance is superior to that of any of its constituent models. The aggregation of models leads to lower overall bias and to a reduction in variance errors~\cite{geron2017hands}.
Bagging and boosting~\cite{russell2016artificial} can improve the performance of ensemble models further. Bagging increases model diversity by selecting data subsets for the training of each model in the ensemble.  After an individual model is created, boosting identifies the samples for which the model performance is deficient, and increases the likelihood of these samples being selected for subsequent model training. AdaBoost~\cite{freund1997decision}, short for Adaptive Boosting, is a widely used boosting algorithm reported to have solved many problems of earlier boosting algorithms~\cite{freund1999short}. Where the training data are imbalanced, the SMOTE boosting algorithm~\cite{chawla2003smoteboost} may be employed.

The selection and optimization of hyperparameters~\cite{geron2017hands} has a significant impact on the performance of  models~\cite{wang2018stealing}. Given the large number of hyperparameters,  tuning them manually is typically unfeasible. 
Automated optimization strategies are employed instead~\cite{hutter2015beyond}, using methods that  include grid search, random search and latin hypercube sampling~\cite{koch2017automated}.  Evolutionary algorithms may also be employed for high-dimensional hyperparameter  spaces~\cite{young2015optimizing}. 
Selecting the most appropriate method for hyperparameter tuning in a given context  and understanding the interaction between hyperparameters and model performance~\cite{lujan2018design}  represent open challenges (ML03 and ML04 in Table ~\ref{tab:MLChallenges}, respectively). 
Furthermore, there are no guarantees that a tuning strategy will continue to be optimal as the model and data on which it is trained evolve. 

When models are updated (or learnt) at run-time, the computational power available may be a limiting factor. While computational costs can be reduced by restricting the complexity of the models selected, this typically leads to a reduction in model performance. As such, a trade-off may be required when computational power is at a premium. For deep learning, which requires significant computational effort, batch normalization~\cite{ioffe2015batch} can lower the computational cost of learning by tackling the problem of vanishing/exploding gradients in the training phase.

\subsubsection{Robust\label{subsubsect:robust}}

Training optimizes models with respect to an objective function using the data in the training set. The aim of the model learning process, however, is to produce a model which generalises to data not present in the training set but which may be encountered in operation. 

Increasing model complexity generally reduces training errors, but noise in the training data may result in overfitting and in a failure of the model to generalise to real-world data. When choosing between competing models one method is then to prefer simple models (Ockham's razor)~\cite{russell2016artificial}. The ability of the model to generalise can also be improved by using $k$-fold cross-validation~\cite{goodfellow2016deep}. This method partitions the training data into $k$ non-overlapping subsets, with $k\!-\!1$ subsets used for training and the remaining subset used for validation. The process is repeated $k$ times, with a different validation subset used each time, and an overall error is calculated as the mean error over the $k$ trials. Other methods to avoid overfitting include gathering more training data, reducing the noise present in the training set, and simplifying the model~\cite{geron2017hands}.  

Data augmentation (Section~\ref{sec:augmentation}) can improve the quality of training data and improve robustness of models~\cite{ko2015audio}. 
Applying transformations to images in the input space may produce models which are robust to changes in the position and orientation of objects in the input space~\cite{geron2017hands} whilst photometric augmentation may increase robustness with respect to lighting and colour~\cite{taylor2017improving}. For models of speech, altering the speed of playback for the training set can increase model robustness~\cite{ko2015audio}.  Identifying the best augmentation methods for a given context can be difficult, and Antoniou et al.~\cite{antoniou2017data} propose an automated augmentation method that uses generative adversarial networks to augment datasets without reference to a contextual setting. 
These methods require the identification of all possible deviations from the training data set, so that  deficiencies can be compensated through augmentation. 
Assuring the completeness of (augmented) data sets has already been identified as an open challenge (DM02 in Table~\ref{tab:DMChallenges}). Even when a data set is complete, the practice of reporting aggregated generalisation errors means that  assessing the impact of each type of deviation on model performance is challenging. Indeed, the complexity of the open environments in which most safety-critical systems operate means that decoupling the effects of different perturbations on model performance remains an open challenge (ML05 in Table~\ref{tab:MLChallenges}).

Regularization methods are intended to reduce a model's generalization error but not its training error~\cite{goodfellow2016deep, russell2016artificial}, e.g., by augmenting the objective function with a term that penalises model complexity. The $\ell_0$, $\ell_1$ or $\ell_2$ norm are commonly used~\cite{Murphy:2012:MLP:2380985}, with the term chosen based on the learning context and model type. The ridge regression~\cite{geron2017hands} method may be used for models with low bias and high variance. This method adds a weighted term to the objective function which aims to keep the weights internal to the model as low as possible.
Early stopping~\cite{prechelt1998early} is a simple method that avoids overfitting by stopping the training if the validation error begins to rise.
For deep neural networks, dropout~\cite{hinton2012improving,srivastava2014dropout} is the most popular regularization method. 
Dropout selects a different subsets of neurons to be ignored at each training step. This makes the model  less reliant on any one neuron, and hence increases its robustness. Dropconnect~\cite{wan2013regularization} 
employs a similar technique to improve the robustness of large networks by setting 
subsets of weights in fully connected layers to zero. 
For image classification tasks, randomly erasing portions of input images can increase the robustness of the generated models~\cite{zhong2017} by ensuring that the model is not overly reliant on any particular subset of the training data. 

Robustness with respect to adversarial perturbations for image classification 
is problematic for deep neural networks, even when the perturbations are imperceptible to humans~\cite{szegedy2013intriguing} or the model is robust to random noise~\cite{fawzi2016robustness}. Whilst initially deemed a consequence of the high non-linearity of neural networks, recent results suggest that the ``success'' of
adversarial examples is due to the low flexibility of classifies, and affects classification models more widely~\cite{fawzi2015fundamental}. Adversarial robustness may therefore be considered as a measure of the \textit{distinguishability} of a classifier. 

Ross and Doshi-Velez~\cite{ross2018improving} introduced a batch normalization method that penalises parameter sensitivity to increase robustness to adversarial examples. This method adds noise to the hidden units of a deep neural network at training time, can have a regularization effect, and sometimes makes dropout unnecessary~\cite{goodfellow2016deep}.
Although regularization can improve model robustness without knowledge of the possible deviations from the training data set, understanding the nature of robustness in a contextually meaningful manner remains an open challenge (ML06 in Table~\ref{tab:MLChallenges}).

\subsubsection{Reusable\label{subsubsect:reusable}}

Machine learning is typically computationally expensive, and repurposing models from related domains can reduce the cost of training new models. Transfer learning~\cite{weiss2016survey} allows for a model  learnt in one domain to be exploited in a second domain, as long as the domains are similar enough so that features learnt in the source domain are applicable to the target domain. Where this is the case, all or part of a model may be transferred to reduce the training cost.

Convolutional neural networks (CNN) are particularly suited for partial model transfer~\cite{geron2017hands} since the convolutional layers encode features in the input space, whilst the fully connected layers encode reasoning based on those features. Thus, a CNN trained on human faces is likely to have feature extraction capabilities to recognise eyes, noses, etc. To train a CNN from scratch for a classifier that considers human faces is wasteful if a CNN for similar tasks already exists. By taking the convolutional layers from a source model and learning a new set of weights for the fully connected set of layers, training times may be significantly reduced~\cite{oquab2014learning, huang2017transfer}.
Similarly, transfer learning has been shown to be effective for random forests~\cite{segev2017learn, sukhija2018supervised}, where subsets of trees can be reused.
More generally, the identification of ``similar'' operational contexts is difficult, and defining a meaningful similarity measure in high-dimensional spaces is an open challenge (ML07 in Table~\ref{tab:MLChallenges}).

Even with significant differences between the source and target domains, an existing model may be valuable. Initialising the parameters of a model to be learnt using values obtained in a similar domain may greatly reduce training times, as shown by the successful use of transfer learning in the classification of sentiments, human activities, software defects, and multi-language texts~\cite{weiss2016survey}. 

Since using pre-existing models as the starting point for a new problem can be so effective, it is important to have access to models previously used to tackle problems in related domains. A growing number of \textit{model zoos}~\cite{geron2017hands} containing such models are being set up by many core learning technology platforms~\cite{modelzoos-github}, as well as by researchers and engineers~\cite{modelzoos-caffe}.

Transfer learning resembles software component reuse, and may allow the reuse of assurance evidence about ML models across domains, as long as the assumptions surrounding the assurance are also transferable between the source and target domains. However, a key aspect in the assurance of components is that they should be reproducible and, at least for complex (deep) ML models, reproducing the learning process is rarely straightforward~\cite{wagstaff2012machine}. Indeed, reproducing ML results requires significant configuration management, which is often overlooked by ML teams~\cite{zaharia2018accelerating}. 

Another reason for caution when adopting third-party model structures, weights and processes is that transfer learning can also transfer failures and faults from the source to the target domain~\cite{gu2017badnets}. Indeed, ensuring that existing models are free from faults is an open challenge (ML08 in Table~\ref{tab:MLChallenges}).

\subsubsection{Interpretable\label{subsubsect:interpretable}}

For many critical domains where assurance is required, it is essential that ML models are interpretable. `Interpretable' and `explainable' are closely related concepts, with `interpretable' used in the ML community and `explainable' preferred in the AI community~\cite{adadi2018peeking}. We use the term `interpretable' when referring to properties of machine learnt models, and `explainable' when  systems features and contexts of use are considered. 

Interpretable models aid assurance by providing evidence which allows for~\cite{lipton2016mythos,adadi2018peeking,630017}: justifying the results provided by a model; supporting the identification and correction of errors;  aiding model improvement; and providing insight with respect to the operational domain. 

The difficulty of providing interpretable models stems from their complexity. By restricting model complexity one can produce models that are intrinsically interpretable; however, this often necessitates a trade-off with model accuracy.

Methods which aid in the production of interpretable models  can be classified by the scope of the explanations they generate. Global methods generate evidence that apply to a whole model, and support design and assurance activities by allowing reasoning about all possible future outcomes for the model. Local methods generate explanations for an individual decision, and may be used to analyse why a particular problem occurred, and to improve the model so future events of this type are avoided.  
Methods can also be classified as model-agnostic and model-specific~\cite{adadi2018peeking}. Model-agnostic methods are mostly applicable post-hoc (after training), and 
include providing natural language explanations~\cite{krening2017learning}, using model visualisations to support understanding~\cite{mahendran2015understanding}, and explaining by example~\cite{adhikari2018example}. Much less common, model-specific methods~\cite{adadi2018peeking} typically provide more detailed explanations, but restrict the users'  choice of model, and therefore are only suited if the limitations of the model(s) they can work with are acceptable.

Despite significant research into interpretable models, there are no global methods providing contextually relevant insights to aid human understanding for complex ML models (ML09 in Table~\ref{tab:MLChallenges}). In addition, although several post-hoc local methods exist, there is no systematic approach to infer global properties of the model from local cases. Without such methods, interpretable models cannot aid structural model improvements and error correction at a global level (ML10 in Table~\ref{tab:MLChallenges}).

\subsection{Summary and Open Challenges}

Table~\ref{tab:MLMethods} summaries the assurance methods applicable during the Model Learning stage. The methods are presented in the order that they are introduced in the preceding discussion, and are matched to the activities with which they are associated and to the desiderata that they support. %From this we can see that there are a substantial number of 
The majority of these methods focus on the performance and robustness of ML models. Model reusability and interpretability are only supported by a few methods that 
% These methods  do little to support the reusable or interpretable desiderata. Indeed there are still relatively few methods for these later desiderata and the methods 
typically restrict the types of model that can be used. This imbalance reflects the different maturity of the research on the four desiderata, with the need for reuse and interpretability arising more prominently after the recent advances in deep learning and increases in the complexity of ML models.

Open challenges for the assurance of the Model Learning stage are presented in Table~\ref{tab:MLChallenges}, organised into categories based on the most relevant desideratum for each challenge. The importance and nature of these challenges have been established earlier in this section. A common theme across many of these challenges is the need for integrating concerns associated with the operating context into the Machine Learning stage (open challenges ML01, ML03, ML05, ML06, ML07). Open challenges also exist in the evaluation of performance of models with respect to multi-objective evaluation criteria (ML02); understanding the link between model performance and hyperparameter selection (ML04) and ensuring that where transfer learning is adopted that existing models are free from errors (ML08). While there has been a great deal of research focused on interpretable models,  methods which apply globally to complex models (ML09) are still lacking. Where local explanations are provided, methods are needed to extract global model properties from them  (ML10).

\begin{table}
	\centering
	\caption{Assurance methods for the Model Learning stage  \label{tab:MLMethods}}
	
	\vspace*{-2mm}
	\begin{footnotesize}
		\def\tabcolsep{1.7pt}
		\sffamily
		\begin{tabular}{L{3.6cm}cccccccc} 
			\toprule
			& \multicolumn{4}{c}{\textbf{Associated activities$^\dagger$}} & \multicolumn{4}{c}{\textbf{Supported desiderata$^\ddagger$}}\\ \cmidrule(l{1pt}r{2pt}){2-5}\cmidrule(l{2pt}r{1pt}){6-9}
			& Model &  Training & Hyperparam.  & Transfer & Performant & Robust & Reusable & Interpretable \\ 
			\textbf{Method} &  Selection &   &  Selection & Learning  &  &  &  & 
			\\ \midrule
			\\[-1.5em]
			\rowcolor{gray!25}
			Use appropriate performance measures~\cite{wagstaff2012machine,flach2019performance}& \smltick & \ding{52} & &  & \ding{72} & \ding{72} & & \\ 
			Statistical tests~\cite{mitchell1997, Murphy:2012:MLP:2380985}& \smltick & \ding{52} & &  & \ding{72} & & & \\ 
			\rowcolor{gray!25}
			Ensemble Learning~\cite{sagi2018ensemble} & \ding{52} & \ding{52} &  & \ding{52} & \ding{72} & \ding{72} &  & \\ 
			Optimise hyperparameters~\cite{hutter2015beyond,young2015optimizing}&  &\ding{52} &\ding{52}  & & \ding{72} & \ding{72} &  & \\ 
			\rowcolor{gray!25}
			Batch Normalization~\cite{ioffe2015batch} &  &\ding{52} &\ding{52}  & & \ding{72} & \ding{72} &  & \\ 
			Prefer simpler models~\cite{russell2016artificial,adhikari2018example} & \ding{52} & \smltick &  &  &\ding{73}  & \ding{72} &  & \ding{73}\\ 
			\rowcolor{gray!25}
			Augment training data &  &\ding{52} &  & & \ding{72} & \ding{72} &  & \\ 
			Regularization methods~\cite{geron2017hands} &  &\ding{52} &\ding{52}  & &  & \ding{72} &  & \\ 
			\rowcolor{gray!25}
			Use early stopping &  &\ding{52} &\ding{52}  & &  & \ding{72} &  & \\
			Use models that intrinsically support reuse~\cite{adadi2018peeking} &\ding{52}  &  &  &\ding{52} &  &  & \ding{72}  & \ding{73} \\
			\rowcolor{gray!25}
			Transfer Learning~\cite{weiss2016survey}& \ding{52}  &   \smltick &  & \ding{52}&  &  & \ding{72} &  \ding{73}\\
			Use model zoos~\cite{geron2017hands} & \ding{52}  &  \smltick&  &\ding{52}  &  &  & \ding{72} & \\
			\rowcolor{gray!25}
			Post-hoc interpretability methods~\cite{krening2017learning,mahendran2015understanding,adhikari2018example} &  & \ding{52} &  & &  &  &   & \ding{72}\\[-0.2em] \bottomrule
			\multicolumn{9}{l}{$^\dagger$\ding{52} = activity that the method is typically used in; \smltick = activity that may use the method}\\
			\multicolumn{9}{l}{$^\ddagger$\ding{72} = desideratum supported by the method; \ding{73} = desideratum partly supported by the method}
		\end{tabular}
	\end{footnotesize}
	\label{tab:ml_astech}
\end{table}

\begin{table}
	\centering
	\caption{Open challenges for the assurance concerns associated with the Model Learning (ML) stage\label{tab:MLChallenges}}
	
	\vspace*{-2mm}
	\begin{footnotesize}
		\sffamily
		\begin{tabular}{L{0.75cm}L{8.0cm}L{3.4cm}}
			\toprule
			\textbf{ID} & \textbf{Open Challenge} & \textbf{Desideratum (Section)}\\ \midrule
			ML01 & Selecting measures which represent operational context  & \multirow{4}{*}{Performant (Section \ref{subsubsect:performant})} \\
			ML02 & Multi-objective performance evaluation at run-time\\
			ML03 & Using operational context to inform hyperparameter-tuning strategies \\
			ML04 & Understanding the impact of hyperparameters on model performance \\ \midrule
			ML05 & Decoupling the effects of perturbations in the input space & \multirow{2}{*}{Robust (Section \ref{subsubsect:robust})} \\
			ML06 & Inferring contextual robustness from evaluation metrics \\ 
			\midrule
			ML07 & Identifying similarity in operational contexts & \multirow{2}{*}{Reusable (Section \ref{subsubsect:reusable})} \\
			ML08 &  Ensuring existing models are free from faults \\ 
			\midrule
			ML09 & Global methods for interpretability in complex models & \multirow{2}{*}{Interpretable (Section \ref{subsubsect:interpretable})} \\ 
			ML10 & Inferring global model properties from local cases  \\ 	\bottomrule
		\end{tabular}
	\end{footnotesize}
	\label{tab:ml_chall}
\end{table}

\section{Model Verification}

The Model Verification stage of the ML lifecycle 
is concerned with the provision of auditable evidence that a model will continue to satisfy its requirements when exposed to inputs which are not present in the training data.

\subsection{Inputs and Outputs}

The key input artefact to this stage is the trained model produced by the Model Learning stage. The key output artefacts are a verified model, and a verification result that provides sufficient information to allow potential users to determine if the model is suitable for its intended application(s).

\subsection{Activities}

\subsubsection{Requirement Encoding}

This activity involves transforming requirements into both tests and  mathematical properties, where the latter can be verified using formal techniques. Requirements encoding requires a knowledge of the application domain, such that the intent which is implicit in the requirements may be encoded as explicit tests and properties. A knowledge of the technology which underpins the model is also required, such that technology-specific issues may be assessed through the creation of appropriate tests and properties. 

\subsubsection{Test-Based Verification}
This activity involves providing test cases (i.e., specially-formed inputs or sequences of inputs) to the trained model and checking the outputs against predefined expected results. A large part of this activity involves an independent examination of properties considered during the Model Learning stage (cf.~Section~\ref{sect:modellearning}), especially those related to the \textsf{Performant} and \textsf{Robust} desiderata. In addition, this activity also considers test completeness, i.e., whether the set of tests exercised the model and covered its input domain sufficiently. The latter objective is directly related to the \textsf{Complete} desideratum from the Data Management stage (cf.~Section~\ref{sect:datamanagement}).

\subsubsection{Formal Verification} 
This activity involves the use of mathematical techniques to provide irrefutable evidence that the model satisfies formally-specified properties derived from its requirements. Counterexamples are typically provided for properties that are violated, and can be used to inform further iterations of activities from the Data Management and Model Learning stages. 

\subsection{Desiderata}
In order to be compelling, the verification results (i.e., the evidence) generated by the Model Verification stage should exhibit the following key properties:
\begin{itemize}
\item \textsf{Comprehensive}---This property is concerned with the ability of Model Verification to cover: (i)~all the requirements and operating conditions associated with the intended use of the model; and (ii)~all the desiderata from the previous stages of the ML lifecycle (e.g., the completeness of the training data, and the robustness of the model).

\item \textsf{Contextually Relevant}---This desideratum considers the extent to which test cases and formally verified properties can be mapped to contextually meaningful aspects of the system that will use the model. For example, for a model used in an autonomous car, robustness with respect to image contrast is less meaningful than robustness to variation in weather conditions.

\item \textsf{Comprehensible}---This property considers the extent to which verification results can be understood by those using them in activities ranging from data preparation and model development to system development and regulatory approval. A clear link should exist between the aim of the Model Verification and the guarantees it provides. Limitations and assumptions should be clearly identified, and results that show requirement violations should convey sufficient information to allow the underlying cause(s) for the violations to be fixed.
\end{itemize}

\subsection{Methods}

\subsubsection{Comprehensive} \label{subsubsect:comprehensive}

Compared to traditional software the dimension and potential testing space of an ML model is typically much larger~\cite{braiek2018testing}. Ensuring that model verification is comprehensive requires a systematic approach to identify faults due to conceptual misunderstandings and faults introduced during the Data Management and Model Learning activities. 

Conceptual misunderstandings may occur during the construction of requirements. They impact both Data Management and Model Learning, and may lead to errors that include: data that are not representative of the intended operational environment; loss functions that do not capture the original intent; and design trade-offs that detrimentally affect performance and robustness when deployed in real-world contexts. 
Independent consideration of requirements is important in traditional software but, it could be argued, it is even more important for ML because the associated workflow includes no formal, traceable hierarchical requirements decomposition \cite{ashmore2017progress}. 
 
Traditional approaches to safety-critical software development distinguish between normal testing and robustness testing \cite{RTCA2011178C}. The former is concerned with typical behaviour, whilst the latter tries to induce undesirable behaviour on the part of the software. In terms of the spaces discussed in Section~\ref{sect:datamanagement}, normal testing tends to focus on the operational domain, $\mathcal{O}$; it can also include aspects of the failure domain, $\mathcal{F}$, and the adversarial domain, $\mathcal{A}$. Conversely, robustness testing utilises the entire input domain, $\mathcal{I}$, including, but not limited to, elements of $\mathcal{F}$ and $\mathcal{A}$. Robustness testing for traditional software is informed by decades of accumulated knowledge on typical errors (e.g., numeric overflow and buffer overruns). Whilst a few typical errors have also been identified for ML (e.g., overfitting and backdoors  \cite{chen2017backdoor}), the knowledge about such errors is limited and rarely accompanied by an understanding of how these errors may be detected and corrected. Developing this knowledge is an open challenge~(challenge MV01 in Table~\ref{tab:MVChallenges}).

Coverage is an important measure for assessing the comprehensiveness of software testing. For traditional software, coverage focuses on the structure of the software. For example, statement coverage or branch coverage can be used as a surrogate for measuring how much of the software's behaviour has been tested. However, measuring ML model coverage in the same way is not informative: achieving high branch coverage for the code that implements a neuron activation function tells little, if anything, about the behaviour of the trained network. For ML, test coverage needs to be considered from the perspectives of both data and model structure. The methods associated with the \textsf{Complete} desiderata from the Data Management stage can inform data coverage. 
In addition, model-related coverage methods have been proposed in recent years \cite{pei2017deepxplore,ma2018deepgauge,sun2018concolic}, although achieving high coverage is generally unfeasible for large models due to the high dimensionality of their input and feature spaces. Traditional software testing employs combinatorial testing to mitigate this problem, and DeepCT~\cite{ma2019deepct} provides combinatorial testing for deep-learning models.

However, whilst these methods provide a means of \emph{measuring} coverage, the benefits of achieving a particular level of coverage are not clear. Put another way, we understand the theoretical value of increased data coverage, but its empirical utility has not been demonstrated. As such, it is impossible to define coverage thresholds that should be achieved. Indeed, it is unclear whether a generic threshold is appropriate, or whether coverage thresholds are inevitably application specific. Consequently, deriving a set of coverage measures that address both data and model structure, and demonstrating their practical utility remains an open challenge (MV02 in Table~\ref{tab:MVChallenges}).

The susceptibility of neural networks to adversarial examples is well known, and can be mitigated using formal verification. Satisfiability modulo theory (SMT) is one method of ensuring local adversarial robustness by providing mathematical guarantees that, for a suitably-sized region around an input-space  point, the same decision will always be returned~\cite{huang2017safety}. The AI$^2$ method~\cite{gehr2018ai2} also assesses regions around a decision point. Abstract interpretation is used to obtain over-app\-roximations of behaviours in convolutional neural networks that utilise the rectified linear unit (ReLU) activation function. Guarantees are then provided with respect to this over-approximation. Both methods are reliant on the assumptions of proximity and smoothness. Proximity concerns the notion that two similar inputs will have similar outputs, while smoothness assumes that the model smoothly transitions between values~\cite{van2017challenges}. However, without an understanding of the model's context, it is difficult to 
ascertain whether two inputs are similar 
(e.g., based on a meaningful distance metric), or to challenge smoothness assumptions when discontinuities are present in the modelled domain. 
In addition, existing ML formal verification methods focus on neural networks rather than addressing the wide variety of ML models. Extending these methods to other types of models represents an open challenge~(MV03 in Table~\ref{tab:MVChallenges}).  

Test-based verification may include use of a simulation to generate test cases. In this case, appropriate confidence needs to be placed in the simulation. For normal testing, the simulation-related concepts discussed in Section~\ref{sect:datamanagement} (e.g., verification and validation) are relevant. If the term `simulation' is interpreted widely, then robustness testing could include simulations that produce pseudo-random inputs (i.e., fuzzing), or simulations that try to invoke certain paths within the model (i.e., guided fuzzing \cite{odena2018tensorfuzz}). In these cases, we need confidence that the `simulation' is working as intended (verification), but we do not need it to be representative of the real world (validation).

While most research effort has focused on the verification of neural networks, there has been some work undertaken to address the problem of verifying other model types. The relative structural simplicity of Random Forests makes them an ideal candidate for systems where verification is required. They too suffer from combinatorial explosion, and so systematic methods are required to provide guarantees of model performance. The VoRF (Verifier of Random Forests) method~\cite{tornblom2018formal} achieves this by partitioning the input domain and exploring all path combinations systematically. 

Last but not least, the model verification must extend to any ML libraries or platforms used in the Model Learning stage. Errors in this software are difficult to identify, as the iterative nature of model training and parameter tuning can mask software implementation errors. Proving the correctness of ML libraries and platforms requires program verification techniques to be applied~\cite{selsam2017developing}. 

\subsubsection{Contextually Relevant} \label{subsubsect:contextuallyrelevant}

Requirement encoding should consider how the tests and formal properties constructed for verification map to context. This is particularly difficult for high-di\-mensional problems such as those tackled using deep neural networks. Verification methods that assess model performance with respect to proximity and smoothness are mathematically provable, but defining regions around a point in space does little to indicate the types of real-world perturbation that can, or cannot, be tolerated by the system (and those that are likely, or unlikely, to occur in reality). As such, mapping requirements to model features is an open challenge~(MV04 in Table~\ref{tab:MVChallenges}). 

Depending on the intended application, Model Verification may need to explicitly consider unwanted bias. In particular, if the context of model use includes a legally-protected characteristic (e.g., age, race or gender) then considering bias is a necessity. As discussed in Section~\ref{subsubsect:relevant}, there are several ways this can be achieved, and an industry-ready toolkit is available \cite{bellamy2018ai}.

Contextually relevant verification methods such as DeepTest~\cite{tian2018deeptest} and DeepRoad~\cite{zhang2018deeproad} have been developed for autonomous driving.  DeepTest employs neuron coverage to guide the generation of tests cases for ML models used in this application domain. Test cases are constructed as contextually relevant transformations of the data set, e.g., by adding synthetic but realistic fog and camera lens distortion to images.
DeepTest leverages principles of metamorphic testing, so that even when the valid output for a set of inputs is unknown it can be inferred from similar cases (e.g., an image with and without camera lens distortion should return the same result). DeepRoad works in a similar way, but generates its contextually relevant images  using a generative adversarial network. 
These methods work for neural networks used in autonomous driving, but developing a general framework for synthesizing test data for other contexts is an open challenge~(MV05 in Table~\ref{tab:MVChallenges}). 

Although adversarial examples are widely used to verify the robustness of neural networks, they typically disregard the semantics and context of the system into which the ML model will be deployed. Semantic adversarial deep learning~\cite{dreossi2018semantic} is a method that avoids this limitation through considering the model context explicitly, first by using input modifications informed by contextual semantics (much like DeepTest and DeepRoad), and second by using system specifications to assess the system-level impact of invalid model outputs. By identifying model errors that lead to system failures, the latter technique aids the model repair and re-design.

The verification of reinforcement learning~\cite{van2017challenges} requires a number of different features to be considered. When Markov decision process (MDP) models of the environment are devised by domain experts, the MDP states are nominally associated with contextually-relevant operating states. As such, systems requirements can be encoded as properties in temporal logics and probabilistic model checkers may be used to provide probabilistic performance guarantees \cite{icaart17}. When these models are learnt from data, it is difficult to map model states to real-world contexts, and constructing meaningful properties is an open challenge (MV06 in Table~\ref{tab:MVChallenges}).

\subsubsection{Comprehensible} \label{subsubsect:comprehensible}

The utility of Model Verification is enhanced if its results provide information that aids the fixing of any errors identified by the test-based and formal verification of ML models. One method that supports the generation of comprehensible verification results is to use contextually relevant testing criteria, as previously discussed. 
Another method is to use counterexample-guided data augmentation~\cite{dreossi2018counterexample}. 
For traditional software, the counterexamples provided by formal verification guide the eradication of errors by pointing to a particular block of code or execution sequence. For ML models, with ground-truth labels by using systematic techniques to cover the modification space. Error tables are then created for all counterexamples, with table columns associated to input features (e.g., car model, environment or brightness for an image classifier used in autonomous driving). The analysis of this table can then provide a comprehensible explanation of failures, e.g., ``The model does not identify white cars driving away from us on forest roads''~\cite{dreossi2018counterexample}. These explanations support further data collection or augmentation in the next iteration of the ML workflow. 
In contrast, providing comprehensible results is much harder for formal verification methods that identify counterexamples based on proximity and smoothness; mapping such counterexamples to guide remedial actions remains an open challenge~(MV07 in Table~\ref{tab:MVChallenges}).

While adding context to training data helps inform how Data Management activities should be modified to improve model performance, no analogous methods exist for adjusting Model Learning activities (e.g., model and hyperparameter selection) in light of verification results. In general, defining a general method for performance improvement based on verification results is an open challenge~(MV08 in Table~\ref{tab:MVChallenges}).

The need for interpretable models is widely accepted; it is also an important part of verification evidence, being comprehensible to people not involved in the ML workflow. However, verifying that a model is interpretable is non-trivial, and a rigorous evaluation of interpretability is necessary. Doshi-Velez and Kim~\cite{doshi2017towards} suggest three possible approaches to achieving this verification: \emph{Application grounded}, which involves placing the explanations into a real application and letting the end user test it; \emph{Human grounded}, which uses lay humans rather than domain experts to test more general forms of explanation; and \emph{Functionally grounded}, which uses formal definitions to evaluate the quality of explanations without human involvement.

\subsection{Summary and Open Challenges}

Table~\ref{tab:MVMethods} summarises the assurance methods that can be applied during the Model Verification stage, listed in the order in which they were introduced earlier in this section. We note that the test-based verification methods outnumber the methods that use formal verification. Furthermore, the test-based methods are model agnostic, while the few formal verification methods that exist are largely restricted to neural networks and, due to their abstract nature, do little to support context or comprehension. 
Finally, the majority of the methods are concerned with the \textsf{Comprehensive} desideratum, while the \textsf{Contextually Relevant} and \textsf{Comprehensible} desiderata are poorly supported.

\begin{table}
	\centering
	\caption{Assurance methods for the Model Verification stage\label{tab:MVMethods} }
	
	\vspace{-2mm}
	\begin{footnotesize}
	
	\def\tabcolsep{1.5pt}
	\sffamily
	\begin{tabular}{L{4.9cm}cccccc} 
	\toprule
        & \multicolumn{3}{c}{\textbf{Associated activities$^\dagger$}} & \multicolumn{3}{c}{\textbf{Supported desiderata$^\ddagger$}}\\ \cmidrule(l{1pt}r{2pt}){2-4}\cmidrule(l{2pt}r{1pt}){5-7}
		 & Requirement & Test-Based & Formal & Compre- &  Contextually & Compre-  \\ 
		\textbf{Method} & Encoding & Verification & Verification & hensive & Relevant & hensible\\ \midrule
		\\[-1.5em]
		\rowcolor{gray!25}
		Independent derivation of test cases & \ding{52} & \smltick & \smltick &  & \ding{72} & \\
		Normal and robustness tests \cite{RTCA2011178C} & \smltick & \ding{52} & & \ding{72} & & \\
		\rowcolor{gray!25}
		Measure data coverage & & \ding{52} & & \ding{72} & \ding{73}&  \\
		Measure model coverage~\cite{pei2017deepxplore,ma2018deepgauge,sun2018concolic} & & \ding{52} & & \ding{72} &\ding{73} &  \\
		\rowcolor{gray!25}
		Guided fuzzing~\cite{odena2018tensorfuzz} & & \ding{52} & & \ding{72} &  &  \\
		Combinatorial Testing~\cite{ma2019deepct} & & \ding{52} & & \ding{72} &  &  \\
		\rowcolor{gray!25}
		SMT solvers~\cite{huang2017safety} & &  & \ding{52} & \ding{72} &  &  \\
		Abstract Interpretation~\cite{gehr2018ai2} & &  & \ding{52} & \ding{72} &  &  \\
   	\rowcolor{gray!25}
		Generate tests via simulation & & \ding{52} & & \ding{72} & \ding{73} & \ding{73} \\
   	Verifier of Random Forests~\cite{tornblom2018formal} & &  & \ding{52} & \ding{72} &  &  \\
 		\rowcolor{gray!25}
		Verification of ML Libraries~\cite{selsam2017developing} & &  & \ding{52} &\ding{72} &  &  \\
		Check for unwanted bias \cite{bellamy2018ai} & & \ding{52} & & & \ding{72} &  \\
 		\rowcolor{gray!25}
 		Use synthetic test data~\cite{tian2018deeptest}&\ding{52} & \ding{52} & &\ding{72} & \ding{72} & \ding{73} \\
 		Use GAN to inform test generation~\cite{zhang2018deeproad}& & \ding{52} & &\ding{72} & \ding{72} &  \\
		\rowcolor{gray!25}
		Incorporate system level semantics~\cite{dreossi2018semantic}&\ding{52} & \ding{52} & &\ding{72} & \ding{72} & \ding{73} \\
		Counterexample-guided data augmentation~\cite{dreossi2018counterexample}& & \ding{52} &   &\ding{72} & \ding{73} & \ding{72} \\
		\rowcolor{gray!25}
		Probabilistic verification~\cite{van2017challenges} & &  & \ding{52} &\ding{72} &  &  \\
		Use confidence levels~\cite{dreossi2018semantic}& & \ding{52} & \smltick  & & \ding{73} & \ding{72} \\
		\rowcolor{gray!25}
		Evaluate interpretability~\cite{doshi2017towards}& & \ding{52} & \ding{52}  & & \ding{72} & \ding{72} \\[-1mm]
\bottomrule
		\multicolumn{7}{l}{$^\dagger$\ding{52} = activity that the method is typically used in; \smltick = activity that may use the method}\\
		\multicolumn{7}{l}{$^\ddagger$\ding{72} = desideratum supported by the method; \ding{73} = desideratum partly supported by the method}
	\end{tabular}
	\end{footnotesize}
	\label{tab:mv_astech}
	
	\vspace*{-2.5mm}
\end{table}

The open challenges for the assurance of the Model Verification stage are presented in Table~\ref{tab:MVChallenges}. 
Much of the existing research for the testing and verification of ML models has focused on neural networks. Providing methods for other ML models remains an open challenge (MV03). A small number of typical errors have been identified but more work is required to develop methods for the detection and prevention of such errors (MV01). Measures of testing coverage are possible for ML models, however, understanding the benefits of a particular coverage remains an open challenge (MV02). Mapping model features to context presents challenges (MV04, MV06) both in the specification of requirements which maintain original intent and in the analysis of complex models. Furthermore, where context is incorporated into synthetic testing, this is achieved on a case by case basis and no general framework for such testing yet exists (MV05). Finally, although formal methods started to appear for the verification of ML models, they return counterexamples that are difficult to comprehend and cannot inform the actions that should be undertaken to improve model performance (MV07, MV08).

\begin{table}
  \centering
		\caption{Open challenges for the assurance concerns associated with the Model Verification (MV) stage\label{tab:MVChallenges}}
		
	\vspace{-2mm}
	\begin{footnotesize}
	\sffamily
  	\begin{tabular}{L{0.75cm}L{8.6cm}L{3cm}}
  	  \toprule
  	  \textbf{ID} & \textbf{Open Challenge} & \textbf{Desideratum (Section)}\\ \midrule
			MV01 & Understanding how to detect and protect against typical errors & Comprehensive \\
			MV02 & Test coverage measures with theoretical and empirical justification & (Section \ref{subsubsect:comprehensive})\\
			MV03 & Formal verification for ML models other than neural networks \\
			 \midrule			
			MV04 & Mapping requirements to model features & Contextually Relevant  \\
			MV05 & General framework for synthetic test generation & (Section \ref{subsubsect:contextuallyrelevant})\\ 
			MV06 & Mapping of model-free reinforcement learning states to real-world contexts \\\midrule
			MV07 & Using proximity and smoothness violations to improve models & Comprehensible  \\
			MV08 & General methods to inform training based on performance failures & (Section \ref{subsubsect:comprehensible})\\ \bottomrule
  	\end{tabular}
  	\end{footnotesize}
		\label{tab:mv_chall}
		
		\vspace*{-2mm}
\end{table}

\section{Model Deployment} \label{sect:modeldeployment}

The aim of the ML workflow is to produce a model to be used as part of a system. How the model is deployed within the system is a key consideration for an assurance argument. The last part of our survey focuses on the assurance of this deployment: we do not cover the assurance of the overall system, which represents a vast scope, well beyond what can be accomplished within this paper. 

\vspace*{-0.5mm}
\subsection{Inputs and Outputs}

The key input artefacts to this stage of the ML lifecycle are a verified model and associated verification evidence. The key output is that model, suitably deployed within a system.

\vspace*{-0.5mm}
\subsection{Activities}

\subsubsection{Integration}
This activity involves integrating the ML model into the wider system architecture. This requires linking system sensors to the model inputs. Likewise, model outputs need to be provided to the wider system. A significant integration-related consideration is protecting the wider system against the effects of the occasional incorrect output from the ML model.

\subsubsection{Monitoring}
This activity considers the following types of monitoring  associated with the deployment of an ML-developed model within a safety-critical system:
\begin{enumerate}
	\item Monitoring the \emph{inputs} provided to the model. This could, for example, involve checking whether inputs are within acceptable bounds before they are provided to the ML model. 
	\item Monitoring the \emph{environment} in which the system is used. This type of monitoring can be used, for example, to check that the observed environment matches any assumptions made during the ML workflow \cite{Aniculaesei2016TowardsTV}.
	\item Monitoring the \emph{internals} of the model. This is useful, for example, to protect against the effects of single event upsets, where environmental effects result in a change of state within a micro-electronic device \cite{taber1993single}.
	\item Monitoring the \emph{output} of the model. This replicates a traditional system safety approach in which a high-integrity monitor is used alongside a lower-integrity item. 
\end{enumerate}

\subsubsection{Updating\label{subsubsect:updating}}
Similarly to software, deployed ML models are expected to require updating during a system's life. This activity relates to managing and implementing these updates. Conceptually it also includes, as a special case, updates that occur as part of online learning (e.g., within the implementation of an RL-based model). However, since they are intimately linked to the model, these considerations are best addressed within the Model Learning stage.

\subsection{Desiderata}

From an assurance perspective, the deployed ML model should exhibit the following key properties:

\begin{itemize}
  \item \textsf{Fit-for-Purpose}---This property recognises that the ML model needs to be fit for the intended purpose within the specific system context. In particular, it is possible for exactly the same model to be fit-for-purpose within one system, but not fit-for-purpose within another. Essentially, this property adopts a model-centric focus.
  \item \textsf{Tolerated}---This property acknowledges that it is typically unreasonable to expect ML models to achieve the same levels of reliability as traditional (hardware or software) components. Consequently, if ML models are to be used within safety-critical systems, the wider system must be able to tolerate the occasional incorrect output from the ML model. 
  \item \textsf{Adaptable}---This property is concerned with the ease with which changes can be made to the deployed ML model. As such, it recognises the inevitability of change within a software system; consequently, it is closely linked to the updating activity described in Section~\ref{subsubsect:updating}.  
\end{itemize}

\subsection{Methods}

This section considers each of the three desiderata in turn. Methods that can be applied during each Model Deployment activity, in order to help achieve the desired property, are discussed.

\subsubsection{Fit-for-Purpose} \label{subsubsect:fitforpurpose}

In order for an ML model to be fit-for-purpose within a given system deployment, there must be confidence that the performance observed during the Model Verification stage is representative of the deployed performance. This confidence could be negatively impacted by changes in computational hardware between the various stages of the ML lifecycle, e.g., different numerical representations can affect accuracy and energy use \cite{hill2018rethinking}. Issues associated with specialised hardware (e.g., custom processors designed for AI applications) may partly be addressed by suitable on-target testing (i.e., testing on the hardware used in the system deployment).

Many safety-critical systems need to operate in real time. For these systems, bounding the worst-case execution time (WCET) of software is important \cite{wilhelm2008worst}. However, the structure of many ML models means that a similar level of computational effort is required, regardless of the input. For example, processing an input through a neural network typically requires the same number of neuron activation functions to be calculated. In these cases, whilst WCET in the deployed context is important, the use of ML techniques introduces no additional complexity into its estimation.

Differences between the inputs received during operational use and those provided during training and verification can result in levels of deployed performance that are very different to those observed during verification. There are several ways these differences can arise:

\begin{enumerate}
	\item Because the training (and verification) data were not sufficiently representative of the operational domain \cite{cieslak2009framework}. This could be a result of inadequate training data (specifically, the $\mathcal{O}$ subset referred to in Section~\ref{sect:datamanagement}), or it could be a natural consequence of a system being used in a wider domain than originally intended. 
	\item As a consequence of failures in the subsystems that provide inputs to the deployed ML model (this relates to the $\mathcal{F}$ subset). Collecting, and responding appropriately to, health management information for relevant subsystems can help protect against this possibility. 
	\item As a result of deliberate actions by an adversary (which relates to the $\mathcal{A}$ subset) \cite{goodfellow2014explaining}, \cite{szegedy2013intriguing}.
	\item Following changes in the underlying process to which the data are related. This could be a consequence of changes in the environment \cite{alaiz2008assessing}. It could also be a consequence of changes in the way that people, or other systems, behave; this is especially pernicious when those changes have arisen as a result of people reacting to the model's behaviour.
\end{enumerate}

The notion of the operational input distribution  being different from that represented by the training  data is referred to as distribution shift \cite{moreno-torres2012distshift}. Most measures for detecting this rely on many operational inputs being available (e.g., \cite{wang2003mining}). A box-based analysis of training data may allow detection  on an input-by-input basis \cite{ashmore2018boxing}. Nevertheless, especially for high-dimensional data \cite{DBLP:journals/corr/abs-1810-11953}, timely detection of distribution shift is an open challenge (MD01 in Table~\ref{tab:sd_chall}).

In order to demonstrate that a deployed model continues to remain fit-for-purpose, there needs to be a way of confirming that the model's internal behaviour is as designed. Equivalently, the provision of some form of built-in test (BIT) is helpful. A partial solution involves re-purposing traditional BIT techniques, including: watchdog timers \cite{pont2002using}, to provide confidence software is still executing; and behavioural monitors \cite{khan2016rigorous},  to provide confidence software is behaving as expected (e.g., it is not claiming an excessive amount of system resources). However, these general techniques need to be supplemented by approaches specifically tailored for ML models \cite{schorn2018efficient}.

For an ML model to be usable within a safety-critical system, its output must be explainable. As discussed earlier, this is closely related to the \textsf{Interpretable} desideratum, discussed in Section~\ref{sect:modellearning}. We also note that the open challenge relating to the global behaviour of a complex model (ML09 in Table~\ref{tab:ml_chall}) is relevant to the Model Deployment stage.

In order to support post-accident, or post-incident, investigations, sufficient information needs to be recorded to allow the ML model's behaviour to be subsequently explained. As a minimum, model inputs should be recorded; if the model's internal state is dynamic, then this should also be recorded. Furthermore, it is very likely that accident, or incident, investigation data will have to be recorded on a continual basis, and in such a way that it will usable after a crash and is protected against inadvertent (or deliberate) alteration. Understanding what information needs to be recorded, at what frequency and for how long it needs to be maintained is an open challenge (MD02 in Table~\ref{tab:sd_chall}).

\subsubsection{Tolerated}  \label{subsubsect:tolerated}

To tolerate occasional incorrect outputs from a deployed ML model, the system needs to do two things. Firstly, it needs to detect when an output is incorrect. Secondly, it needs to replace the incorrect output with a suitable value to allow system processing activities to continue.

 An ML model may produce an incorrect output when used outside the intended operational environment. This could be detected by monitoring for distribution shift, as indicated in the preceding section, possibly alongside monitoring the environment.  An incorrect output may also be produced if the model is provided with inappropriate inputs. Again, this could be detected by monitoring for distribution shift or by monitoring the health of the system components that provide inputs to the model. More generally, a minimum equipment list should be defined. This list should describe the equipment that must be present and functioning correctly to allow safe use of the system \cite{munro2003analysis}. This approach can also protect the deployed ML model against system-level changes that would inadvertently affect its performance.

It may also be possible for the ML model to calculate its own `measure of confidence', which could be used to support the detection of an incorrect output. The intuitive certainty measure (ICM) has been proposed \cite{van2018icm}, but this requires a distance metric to be defined on the input domain, which can be difficult. More generally, deriving an appropriate measure of confidence is an open challenge (MD03 in Table~\ref{tab:sd_chall}).

Another way of detecting incorrect outputs involves comparing them with `reasonable' values. This could, for example, take the form of introducing a simple monitor, acting directly on the output provided by the ML model \cite{bogdiukiewicz2017formal}. If the monitor detects an invalid output then the model is re-run (with the same input, if the model is non-deterministic, or with a different input). Defining a monitor that protects safety is possible \cite{machin2018smof}, but providing sufficient protection yet still allowing the ML model sufficient freedom in behaviour, so that the benefits of using an ML-based approach can be realised, is an open challenge (MD04 in Table~\ref{tab:sd_chall}).

The difficulty with defining a monitor may be overcome by using multiple, `independent' ML models, along with an `aggregator' that combines their multiple outputs into a single output . This can be viewed as an ML-based implementation of the concept of n-version programming \cite{chen1995n}. The approach has some similarity to ensemble learning \cite{sagi2018ensemble}, but its motivation is different: ensemble learning aims to improve performance in a general sense, while using multiple, independent models as part of a system architecture aims to protect against the consequences of a single model occasionally producing an incorrect output. Whilst this approach may have value, it is not clear how much independence can be achieved, especially if models are trained from data that have a common generative model \cite{fawzi2018adversarial}. Consequently, understanding the level of independence that can be introduced into models trained on the same data is an open challenge (MD05 in Table~\ref{tab:sd_chall}).

If an incorrect output is detected then, as noted above, a replacement value needs to be provided to the rest of the system. This could be provided by a software component developed and verified using traditional techniques \cite{caseley2016claims, heitmeyer1996automated}.  Alternatively, a fixed `safe' value or the `last-good' output provided by the ML model could be used. In this approach, a safety switch monitors the output from the ML model. If this is invalid then the switch is thrown and the output from the `alternative' approach is used instead. This assumes that an invalid output can be detected and, furthermore, that a substitute, potentially suboptimal, output can be provided in such cases.  

The monitor, aggregator and switch model-deployment architectures could readily accommodate human interaction. For example, a human could play the role of a monitor, or that allocated to traditional software (e.g., when an autonomous vehicle hands control back to a human driver).

\subsubsection{Adaptable} \label{subsubsect:adaptable}

Like all software, a deployed ML model would be expected to change during the lifetime of the system in which it is deployed. Indeed, the nature of ML, especially the possibility of operational systems capturing data that can be used to train subsequent versions of a model, suggests that ML models may change more rapidly than is the case for traditional software.

A key consideration when allowing ML models to be updated is the management of the change from an old version of a model to a new version. Several approaches can be used for this purpose:

\begin{enumerate}
	\item Placing the system in a `safe state' for the duration of the update process. In the case of an autonomous vehicle, this state could be stationary, with the parking brake applied, with no occupants and with all doors locked. In addition, updates could be restricted to certain geographic locations (e.g., the owner's driveway or the supplier's service area).
	\item If it is not feasible, or desirable, for the system to be put into a safe state then an alternative is for the system to run two identical channels, one of which is `live' and the other of which is a `backup'. The live model can be used whilst the backup is updated. Once the update is complete, the backup can become live and the other channel can be updated.
	\item Another alternative is to use an approach deliberately designed to enable run-time code replacement (or `hot code loading'). This functionality is available, e.g., within Erlang \cite{carlsson2000core}.
\end{enumerate} 

ML model updating resembles the use of field-loadable software in the aerospace domain \cite{RTCA2011178C}. As such, several considerations are common to both activities: detecting corrupted or partially loaded software; checking compatibility; and preventing inadvertent triggering of the loading function. Approaches for protecting against corrupted updates should cover inadvertent data changes  and deliberate attacks aiming to circumvent this protection \cite{meyer2013sok}. 

In the common scenario where multiple instances of the same system have be deployed (e.g., when a manufacturer sells many units of an autonomous vehicle or medical diagnosis system) updates need to be managed at the ``fleet'' level. There may, for example, be a desire to gradually roll out an update so that its effects can be measured, taking due consideration of any ethical issues associated with such an approach. More generally, there is a need to monitor and control fleet-wide diversity \cite{ashmore2019rethinking}. Understanding how best to do this is an open challenge (MD06 in Table~\ref{tab:sd_chall}).

\subsection{Summary and Open Challenges}

Table~\ref{tab:sd_astech} summarises assurance methods associated with the Model Deployment stage, matched to the associated activities and desiderata. The table shows that only two methods support the activity of updating an ML model. This may reflect the current state of the market for autonomous systems: there are few, if any, cases where a manufacturer has a large number of such systems in operational use. Given the link between the updating activity and the \textsf{Adaptable} desideratum, similar reasons may explain the lack of methods to support an adaptable system deployment.

\begin{table}
	\centering
	\caption{Assurance methods for the Model Deployment stage}
	
	\vspace*{-2mm}
	\begin{footnotesize}
	\def\tabcolsep{1.7pt}
	\sffamily
	\begin{tabular}{L{5.3cm}cccccc} 
	\toprule
        & \multicolumn{3}{c}{\textbf{Associated activities$^\dagger$}} & \multicolumn{3}{c}{\textbf{Supported desiderata$^\ddagger$}}\\ \cmidrule(l{1pt}r{2pt}){2-4}\cmidrule(l{2pt}r{1pt}){5-7}
		\textbf{Method} & Integration & Monitoring & Updating & Fit-for-Purpose & Tolerated & Adaptable \\ \midrule
		\\[-1.5em]
   	    \rowcolor{gray!25}
		Use the same numerical precision for training and operation & \ding{52} & & & \ding{72} & & \\
		Establish WCET \cite{wilhelm2008worst} & \ding{52} & & & \ding{72} & \ding{73} & \\
   	    \rowcolor{gray!25}
		Monitor for distribution shift \cite{moreno-torres2012distshift}, \cite{ashmore2018boxing} & \smltick & \ding{52} & & \ding{72} & \ding{72} & \\
		Implement general BIT \cite{pont2002using}, \cite{khan2016rigorous}, \cite{schorn2018efficient} & \ding{52} & \ding{52} & & \ding{72} & \ding{72} & \\
   	    \rowcolor{gray!25}
		Explain an individual output \cite{ribeiro2016should} & \smltick & \ding{52} & & \ding{72} & & \\
		Record information for post-accident (or post-incident) investigation & \ding{52} & & & \ding{72} & & \\
   	    \rowcolor{gray!25}
		Monitor the environment \cite {Aniculaesei2016TowardsTV} & & \ding{52} & & \ding{72} & \ding{72} & \\
		Monitor health of input-providing subsystems & & \ding{52} & & \ding{72} & \ding{72} & \\
   	    \rowcolor{gray!25}
		Provide a confidence measure \cite{van2018icm} & \ding{52} & \ding{52} & & & \ding{72} & \\
		Use an architecture that tolerates incorrect outputs \cite{bogdiukiewicz2017formal}, \cite{caseley2016claims}, \cite{chen1995n} & & \ding{52} & & & \ding{72} & \\
   	    \rowcolor{gray!25}
		Manage the update process \cite{RTCA2011178C} & & \smltick & \ding{52} & & & \ding{72} \\
		Control fleet-wide diversity \cite{ashmore2019rethinking} & & & \ding{52} & & & \ding{72} \\ \bottomrule
		\multicolumn{7}{l}{$^\dagger$\ding{52} = activity that the method is typically used in; \smltick = activity that may use the method}\\
		\multicolumn{7}{l}{$^\ddagger$\ding{72} = desideratum supported by the method; \ding{73} = desideratum partly supported by the method}
	\end{tabular}
	\end{footnotesize}
	\label{tab:sd_astech}
	
	\vspace*{-2mm}
\end{table}

The open challenges associated with the System Deployment stage (Table~\ref{tab:sd_chall}) include: concerns that extend to the Model Learning and Model Verification stages, e.g., providing measures of confidence (MD03); concerns that relate to system architectures, e.g., detecting distribution shift (MD01), supporting incident investigations (MD02), providing suitably flexible monitors (MD04) and understanding independence (MD05); and concerns that apply to system ``fleets'' (MD06).

\begin{table}
  \centering
		\caption{Open challenges for the assurance concerns associated with the Model Deployment (MD) stage}
		
	\vspace*{-2mm}
	\begin{footnotesize}
	\sffamily
  	\begin{tabular}{L{0.75cm}L{9.2cm}L{2.9cm}}
  	  \toprule
  	  \textbf{ID} & \textbf{Open Challenge} & \textbf{Desideratum (Section)}\\ \midrule
			MD01 & Timely detection of distribution shift, especially for high-dimensional data sets & \multirow{2}{*}{\shortstack[l]{Fit-for-Purpose\\ (Section \ref{subsubsect:fitforpurpose}\texttt{})}} \\
			MD02 & Information recording to support accident or incident investigation \\ \midrule
			MD03 & Providing a suitable measure of confidence in ML model output & \multirow{3}{*}{Tolerated (Section \ref{subsubsect:tolerated})} \\
			MD04 & Defining suitably flexible safety monitors \\
			MD05 & Understanding the level of independence that can be introduced into models trained on the same data \\ \midrule		
			MD06 & Monitoring and controlling fleet-wide diversity & Adaptable (Section \ref{subsubsect:adaptable}) \\ \bottomrule
  	\end{tabular}
  	\end{footnotesize}
		\label{tab:sd_chall}
		
		\vspace*{-2mm}
\end{table}

%!TEX root = ../main.tex

\section{Conclusion \label{sect:conclusion}}

\vspace*{-1mm}
Recent advances in machine learning underpin the development of many successful systems. ML technology is increasingly at the core of sophisticated functionality provided by smart devices, household appliances and online services, often unbeknownst to their users. Despite the diversity of these ML applications, they share a common characteristic: none is safety critical. Extending the success of machine learning to safety-critical systems holds great potential for application domains ranging from healthcare and transportation to manufacturing, but requires the assurance of the ML models deployed within such systems. Our paper explained that this assurance must cover all stages of the ML lifecycle, defined assurance desiderata for each such stage, surveyed the methods available to achieve these desiderata, and highlighted remaining open challenges.

For the Data Management stage, our survey shows that a wide range of data collection, preprocessing, augmentation and analysis methods can help ensure that ML training and verification data sets are \textsf{Relevant}, \textsf{Complete}, \textsf{Balanced} and \textsf{Accurate}. Nevertheless, further research is required to devise methods capable of demonstrating that these data are sufficiently secure, fit-for-purpose and, when simulation is used to synthesise data, that simulations are suitably realistic.

The Model Learning stage has been the focus of tremendous research effort, and a vast array of model selection and learning methods are available to support the development of \textsf{Performant} and \textsf{Robust} ML models. In contrast, there is a significant need for additional hyperparameter selection and transfer learning methods, and for research into ensuring that ML models are \textsf{Reusable} and \textsf{Interpretable}, in particular through providing context-relevant explanations of behaviour. 

Assurance concerns associated with the Model Verification stage are addressed by numerous test-based verification methods and by a small repertoire of recently introduced formal verification methods. The verification results provided by these methods are often \textsf{Comprehensive} (for the ML model aspects being verified) and, in some cases, \textsf{Contextually Relevant}. However, there are currently insufficient methods capable of encoding the requirements of the model being verified into suitable tests and formally verifiable properties. Furthermore, ensuring that verification results are \textsf{Comprehensible} is still very challenging.

The integration and monitoring activities from the Model Deployment stage are supported by a sizeable set of methods that can help address the \textsf{Fit-for-Purpose} and \textsf{Tolerated} desiderata of deployed ML models. These methods are often inspired by analogous methods for the integration and monitoring of software  components developed using traditional engineering approaches. 
In contrast, ML model updating using data collected during operation has no clear software engineering counterpart. As such, model updating methods are scarce and typically unable to provide the assurance needed to deploy ML models that are \textsf{Adaptable} within safety-critical systems.

This brief summary shows that considerable research is still needed to address outstanding assurance concerns associated with every stage of the ML lifecycle. In general, using ML components within safety-critical systems poses numerous open challenges. At the same time, the research required to address these challenges can build on a promising combination of rigorous methods developed by several decades of sustained advances in machine learning, in software and systems engineering, and in assurance development. 

\vspace*{-1mm}
\section*{Acknowledgements}

\vspace*{-1mm}
This work was partly funded by the Assuring Autonomy International Programme.

\vspace*{-1mm}
\citestyle{acmnumeric}
\bibliographystyle{ACM-Reference-Format}
\bibliography{ms}

\end{document}